\documentclass[12pt]{article}
\usepackage{amsmath}
\usepackage{graphicx}
\usepackage{enumerate}
\usepackage{natbib}
\usepackage{url} 

\usepackage{tikz}
\usepackage{graphicx}
\usepackage{epsfig}
\usepackage{amsmath}
\usepackage{hyperref}

\usepackage{amsmath}
\usepackage{algorithm}
\usepackage{algpseudocode}

\usepackage{subcaption}

\usepackage{xr}
\usepackage{listofitems} 
\usetikzlibrary{arrows.meta} 
\usepackage[outline]{contour} 
\contourlength{1.4pt}

\usepackage{listings}

\usepackage{xcolor}
\colorlet{myred}{red!80!black}
\colorlet{myblue}{blue!80!black}
\colorlet{mygreen}{green!60!black}
\colorlet{myorange}{orange!70!red!60!black}
\colorlet{mydarkred}{red!30!black}
\colorlet{mydarkblue}{blue!40!black}
\colorlet{mydarkgreen}{green!30!black}

\usepackage{algpseudocode}
\usepackage{multirow}
\usepackage{latexsym, amsbsy, amssymb}
\usepackage{verbatim}
\usepackage{color}

\def\var{{\mathrm{var}}}

\def\nano{\scriptscriptstyle}
\def\real{\mathbb R}
\def\natural{\mathbb N}

\def\var{\mathrm{var}}

\def\nano{\scriptscriptstyle}
\def\inv{\hi{\nano -1}}

\def\nano{\scriptscriptstyle}

\def\ka{\kappa}

\def\ali{&\,}

\def\hii#1{\hi{\mbox{\tiny {(\uppercase{#1})}}}}

\def\b1{\mathbf{1}}

\def\rom#1{ 
  \textup{\uppercase\expandafter{\romannumeral#1}}%
}

\newlength\myindent
\allowdisplaybreaks

\def\boe{\begin{enumerate}}
\def\eoe{\end{enumerate}}

\newtheorem{corollary}{{\bf Corollary}}
\newtheorem{assumption}{{\bf Assumption}}
\newtheorem{definition}{{\bf Definition}}
\newtheorem{theorem}{{\bf Theorem}}

\newcommand\ca[1]{{\cal{#1}}}
\newcommand\lo[1]{_{\nano{#1}}}
\newcommand\hi[1]{^{\nano{#1}}}

\def\argmin{\mathrm{argmin}}

\def\L{{\cal L}}

\def\trans{^{\mbox{\tiny{\sf  T}}}}

\def\inv{^{\mbox{\tiny $-1$}}}

\def\var{\mathrm{var}}

\newcommand{\indep}{\;\, \rule[0em]{.03em}{.65em} \hspace{-.45em}
\rule[-.05em]{.65em}{.03em} \hspace{-.45em}
\rule[0em]{.03em}{.65em}\;\,}

\def\trans{^{\mbox{\tiny{\sf T}}}}

\def\ali{&\,}

\def\nano{\scriptscriptstyle}

\def\real{{\mathbb R}}

\def\ka{\kappa}

\def\L2T{L \lo 2 (T)}
\def\L2TX{L \lo 2 (T\lo X)}
\def\L2TX{L \lo 2 (T\lo Y)}

\def\ali{&\,}

\def\ali{&\,}

\def\eod{\end{document}}

\def\A{{\bf A}}

\def\B{{\bf B}}
\def\b{{\bf b}}

\def\f{{\bf f}}

\def\g{{\bf g}}

\def\h{{\bf h}}

\def\k{{\bf k}}
\def\L{{\bf L}}

\def\s{{\bf s}}

\def\t{{\bf t}}
\def\T{{\bf T}}
\def\u{{\bf u}}
\def\m{{\bf m}}
\def\v{{\bf v}}
\def\W{{\bf W}}
\def\T{{\bf T}}

\def\X{{\bf X}}

\def\x{{\bf x}}
\def\Y{{\bf Y}}
\def\y{{\bf y}}

\def\z{{\bf z}}

\def\0{{\bf 0}}
\def\1{{\bf 1}}

\def\nn{_{\mathrm{NN}}}
\def\benn{_{\mathrm{BENN}}}

\newcommand{\blind}{1}

\begin{document}

\bibliographystyle{apalike}

\def\spacingset#1{\renewcommand{\baselinestretch}%
{#1}\small\normalsize} \spacingset{1}


\if1\blind
{
  \title{\bf Belted and Ensembled Neural Network for Linear and Nonlinear Sufficient Dimension Reduction}
  \author{Yin Tang and Bing Li\thanks{
    Bing Li’s research is partly supported by NSF DMS-2210775 and NIH 1 R01 GM152812-01.}\hspace{.2cm}\\
    Department of Statistics, The Pennsylvania State University}
  \maketitle
} \fi

\if0\blind
{
  \bigskip
  \bigskip
  \bigskip
  \begin{center}
    {\LARGE\bf Belted and Ensembled Neural Network for Linear and Nonlinear Sufficient Dimension Reduction}
\end{center}
  \medskip
} \fi

\bigskip
\begin{abstract}
We introduce a unified,  flexible, and easy-to-implement  framework of sufficient dimension reduction that can accommodate both linear and nonlinear dimension reduction, and both the conditional distribution and the conditional mean as the targets of estimation.  This unified framework is achieved by a specially structured neural network --- the Belted and Ensembled Neural Network (BENN) --- that consists of a narrow latent layer, which we call the belt,   and a family of transformations of the response, which we call the ensemble.  By strategically placing the belt at different layers of the neural network, we can achieve linear or nonlinear sufficient dimension reduction, and by choosing the appropriate transformation families, we can achieve dimension reduction for the conditional distribution or the conditional mean. 
Moreover, thanks to the advantage of the neural network,  the method is very fast to compute, overcoming a computation bottleneck of the traditional sufficient dimension reduction estimators, which involves the inversion of a matrix of dimension either $p$ or $n$. We develop the algorithm and convergence rate of our method, compare it with existing sufficient dimension reduction methods, and apply it to two data examples. 
\end{abstract}

\noindent%
\vfill

\newpage
\spacingset{1.9} 

\section{Introduction}

Sufficient dimension reduction (SDR) is a methodology that extracts a low-dimensional sufficient predictor from the high-dimensional observed predictor. It is a powerful tool for data visualization, regression diagnostics, enhancement of regression accuracy, and assessment of conditional independence. See, for example,   \cite{Li1991} and \cite{Cook1994},  \cite{li2018sufficient}, and \cite{li2024sufficient}. SDR has been developed into many branches, but the  {four most important} branches can be characterized as follows. According to the target of estimation, there are methods that target the conditional distribution (\cite{Li1991}, \cite{Cook1991}) and those that target the conditional mean (\cite{Cook2002,cook-li-2004}, \cite{xia-tong-li-zhu-2002}). According to the nature of the extracted predictor, there are linear and nonlinear SDR methods (\cite{lee2013} and \cite{li2017nonlinear}). A computational bottleneck in scaling up these methods for big data is that they involve   inversion  of matrices with dimensions  equal to the sample size or  the dimension of the predictor,  which can be very slow for large and/or high-dimensional data. In this paper we propose a deep-learning-based SDR methodology that is flexible enough to accommodate all of the above four branches, that is fast to compute because it avoids the inversion of large matrices, and that is highly accurate due to the richness of the deep learning family.

To properly motivate our method, we first give an overview of the four frameworks mentioned above. 
Let $\X$ be a random vector of covariates in $\real \hi p$, and $Y$ be the response in $\real$. The general  goal is to find a transformation $\f: \real \hi p \to \real \hi d$, where $d<p$, such that 
\begin{align}\label{eq:nonlinear-sdr-cs}
   Y \indep \X | \f (\X);  
\end{align}
that is, $Y$ and $\X$ are independent given $\f (\X)$.  The function  $\f (\X)$ is called the  sufficient predictor of $Y$ as it contains all the information in $\X$ about  $Y$.

When $\f$ in (\ref{eq:nonlinear-sdr-cs}) is a linear function,  the dimension reduction problem becomes 
\begin{align}\label{eq:linear-sdr-cs}
    Y \indep \X | \B \trans \X, 
\end{align}
which is known as  the linear SDR problem.  See  \cite{Li1991} and \cite{Cook1994}. 
The identifiable object in (\ref{eq:linear-sdr-cs}) is the subspace of $\real \hi p$ spanned by the columns of $\B$, since the relation (\ref{eq:linear-sdr-cs}) is not affected if we right multiply $\B$ by a nonsingular square matrix. The smallest such space is called the {\em central subspace}, denoted by $\ca S \lo {Y | \X}$. This is the objective of estimation for linear SDR. Many methods are developed for linear SDR, which are summarized, for example, in \cite{li2018sufficient,ma2013review}. 
When $\f$ is a nonlinear function, problem (\ref{eq:nonlinear-sdr-cs}) is the nonlinear SDR problem and the identifiable (infinite-dimensional) object  is the $\sigma$-field generated by $\f (\X)$. The smallest such $\sigma$-field is called  the {\em central $\sigma$-field}, which  is the objective of estimation for nonlinear SDR.   \cite{lee2013} and \cite{li2017nonlinear} developed estimation procedures for nonlinear SDR based on the reproducing-kernel Hilbert spaces (RKHS).

A closely related SDR problem is one in which conditional mean $E(Y|\X)$ is of primary interest. In  this setting, we assume
\begin{align}\label{eq:nonlinear cms}
    E ( Y | \X ) = E [ Y | \f (\X)]. 
\end{align}
When $\f$ is a linear function,  {the above problem \eqref{eq:nonlinear cms} becomes }
\begin{align}\label{eq:linear-sdr-cms}
    E(Y| \X) = E(Y | \B \trans \X).
\end{align}
{This is} the SDR problem considered by  \cite{Cook2002,cook-li-2004}. As in  (\ref{eq:linear-sdr-cms}), this relation is determined by the column space of $\B$, and the smallest such space is called the {\em central mean subspace}. The  forward regression methods proposed by \cite{xia-tong-li-zhu-2002} in fact target  this problem.   It is also useful to consider the problem (\ref{eq:linear-sdr-cms}) when $\f$ is a nonlinear function. The smallest $\sigma$-field that satisfies  (\ref{eq:nonlinear cms}) is called  the {\em central mean $\sigma$-field}. Note that problem (\ref{eq:nonlinear cms}) is meaningful only when the response is a random vector $\Y$ because, when the response is a scalar $Y$, we can always set $\f (\X)= E (Y|\X)$ to satisfy (\ref{eq:nonlinear cms}).

One can bridge the difference between the SDR problems (\ref{eq:nonlinear-sdr-cs}) and (\ref{eq:nonlinear cms}) by the use of an {\em ensemble}. Suppose $\{ g(\cdot, t): t \in I\}$, where $I$ is a subset of $\real$,  is a family of transformations of $Y$ that uniquely determines the distribution of $Y$. Then 
\begin{align*}
    Y \indep \X | \f (\X) \ \Leftrightarrow \ 
    E [ g (Y, t) | \X ] = E [ g (Y, t) | \f (\X)] \quad \mbox{for all $t \in I$}. 
\end{align*}
Such a class of functions is called a probability determining class, and we call it an {\em ensemble} in our context. 
The above equivalence means that we can recover the central subspace (or  $\sigma$-field) by  estimating the central mean subspace (or sub-$\sigma$-field) for every member of the ensemble. This idea was first used by  \cite{zhu2006fourier}, who propose  a Fourier transform method to estimate the central mean space in linear SDR problems.
\cite{yinli2011} formally introduces the  ensemble estimator, and proposes  various ensembles to turn any method for estimating the central mean subspace into one that estimate the central subspace. 
See also \cite{li2008projective}, \cite{fertl2022ensemble}, and   \cite{zeng2010integral}.

From the above discussion,  we need the following ingredients to conduct linear or nonlinear sufficient dimension reduction: 
\begin{enumerate}
  \item a family of functions to model the sufficient predictor $\f (\X) \in \real \hi d$; 
  \vspace{-.1in}
  \item a family of functions of $Y$ (the ensemble) to characterize the conditional distribution of $Y$ given $\X$; 
  \vspace{-.1in}
  \item a regression operation that links the above two families.  
\end{enumerate}
For linear SDR, the first family is the linear family $\B\trans \X$; the second  family is any ensemble; the regression operation can be the forward regression or inverse regression. For RKHS-based nonlinear SDR procedures mentioned earlier, both the first and the second families are RKHS's, and the regression operation is implemented by a linear operator known as the {\em regression operator}.

In this paper, we introduce a deep-learning framework (see, for example, \cite{Goodfellow-et-al-2016} and 
\cite{yuan2020deep}) for linear and nonlinear SDR,  where the first family  is implemented by   neural network, the ensemble is any distribution-determining class, and the regression operation that links the two is also implemented by neural network.  Specifically, at the population level, we minimize 
\begin{align}\label{eq:most general form}
E \|   \g (Y) - ( {\h\nn} \circ {\f\nn}) (\X) \| \hi 2, 
\end{align}
where $\g(Y)$ is a vector-valued function of $Y$ that plays the role of the ensemble, ${\f\nn}$ is a neural network mapping from $\real \hi p$ to $\real \hi d$ used to model the sufficient predictor, and ${\h\nn}$ is another neural network that maps the sufficient predictor into the ensemble space. We call ${\f\nn}$ the dimension reduction neural network, and ${\h\nn}$ the ensemble neural network. Since   the output of ${\f\nn}$ is of dimension  $d$, narrower than the other layers, we call   ${\h\nn} \circ {\f\nn}$ the {\em Belted and Ensembled Neural Network} (BENN).

We can see immediately the neural network framework is flexible enough to accommodate all four settings mentioned earlier. If we place the belt as the first hidden layer {without an activation function}, then the above objective function becomes 
$
E \|   \g (Y) - {\h\nn}   (\B \trans \X) \| \hi 2,  
$ 
which corresponds to the linear SDR problem (\ref{eq:linear-sdr-cs}). If we place the belt in the first hidden layer  {as above} and {further} take $g$ to be the identity mapping $g (Y) = Y$, then (\ref{eq:most general form}) becomes  
$
E [  Y - {h\nn}   (\B \trans \X) ] \hi 2,  
$ 
which corresponds to conditional mean linear SDR problem (\ref{eq:linear-sdr-cms}). If we place the belt in the middle of the neural network, then the objective function  (\ref{eq:most general form}) takes its original form, which  corresponds to the nonlinear SDR problem (\ref{eq:nonlinear-sdr-cs}). If we place the belt in the middle and take $g$ to be the identity mapping, then (\ref{eq:most general form}) becomes 
$
E [  Y - {h\nn} \circ {\f\nn} (\X) ] \hi 2,  
$ 
which 
corresponds to the conditional mean nonlinear SDR problem (\ref{eq:nonlinear cms}).    

Several recent papers have pioneered   the  neural network approaches to sufficient dimension reduction. \cite{kapla2022fusing} proposes a neural-network-based linear SDR method that corresponds exactly to the conditional mean linear SDR described above. Thus, our proposed BENN framework can be regarded as an extension of their approach. \cite{liang2022nonlinear} proposes a stochastic neural network (StoNet) for nonlinear SDR. Each hidden layer of   StoNet involves auxiliary Gaussian noises, and the output of the last hidden layer is used as the sufficient predictor. Since the sufficient predictor produced by  the StoNet depends on the random errors in each hidden layer,  it is not a deterministic function of $\X$ (or deterministic $\sigma$-field) as postulated in the original SDR problem (\ref{eq:nonlinear-sdr-cs}). \cite{sun2022kernel} further generalizes the StoNet to a kernel-expanded stochastic neural network (K-StoNet).  \cite{huang2024deep} introduces  a nonlinear SDR method based on neural networks by maximizing the distance covariance between the sufficient predictor $\f(\X)$ and the response $Y$, and add a penalty term for identifiability. For further related developments, see \cite{chen2024deep}.

The rest of the paper is organized as follows. In Section \ref{sec:concept},  we introduce our method at the population level, describe its mathematical structure and the intuitions behind it. In Section \ref{sec:varieties and precursors}, we discuss the various important special cases of our framework and its relations with some existing methods. In Section \ref{sec:implementation}, we  develop the numerical procedure  to implement our method at the sample level. In Section \ref{sec:rate}, we derive the convergence rate of the proposed estimator, and in Section \ref{sec:rate 1} in the Supplementary Material, we show that  this rate is faster than that of the neural network regression without dimension reduction.     In Section \ref{sec:simulation}, we compare by simulation our proposed method with several existing linear and nonlinear SDR methods based on neural networks or RKHS. In Section \ref{sec:application}, we   apply our method to a data application. To save space, all proofs, some additional simulations, and another data application  are presented in a separate  Supplementary Material.

\section{Sufficient dimension reduction via BENN}\label{sec:concept}

\subsection{General sufficient dimension reduction  through ensembles}

Let $(\Omega, \ca F, P)$ be a probability space, $(\Omega \lo \X, \ca F \lo \X)$ and $(\Omega \lo Y, \ca F \lo Y)$ be measurable spaces, where $\Omega \lo \X \subseteq \real \hi p$ and $\Omega \lo Y \subseteq \real$, and $\ca F \lo \X$ and $\ca F \lo Y$ are Borel $\sigma$-fields. Let $\X: \Omega \to \Omega \lo \X$ and $Y: \Omega \to \Omega \lo Y $ be random variables measurable with respect to $\ca F / \ca F \lo \X$ and $\ca F / \ca F \lo Y$, respectively. 
As is stated in the Introduction, in sufficient dimension reduction, we postulate the conditional independence (\ref{eq:nonlinear-sdr-cs}) and, under this hypothesis, seek to estimate the statistic $\f (\X)$, or any random vector that has a one-to-one relation with $\f (\X)$. We call (\ref{eq:nonlinear-sdr-cs}) the general sufficient dimension reduction problem as $\f$ can be a linear or a nonlinear functions of $\X$. When $\f$ is linear,  (\ref{eq:nonlinear-sdr-cs}) is called the linear SDR problem; when $\f$ is nonlinear,   (\ref{eq:nonlinear-sdr-cs}) is called the nonlinear SDR problem.  The condition (\ref{eq:nonlinear-sdr-cs}) can be restated in terms of the characteristic family--or the ensemble--as we define below.

\begin{definition}\label{def:characteristic}
    We say that a function class $\ca G$ characterizes the distributions on $\Omega \lo Y$ if,   for any two random variables $Y \lo 1$ and $Y \lo 2$ taking values in $\Omega \lo Y$, $E[g(Y \lo 1)] = E[g(Y \lo 2)]$ for all $g \in \ca G$ implies that $Y \lo 1 \overset{\ca D}{=} Y \lo 2$, where $\overset{\ca D}{=}$ indicates equal in distribution. Such a function class is called an ensemble. 
\end{definition}

To be more specific, in this paper we consider the parametric family of functions
\begin{align}\label{eq:ensemble}
    \ca G = \{ g(\cdot, t) : t \in I \},
\end{align}
where $I$ is a subset of $\real$. Many parametric families are characteristic. For example,  when $Y$ is a scalar, 
\begin{align}\label{eq:ensemble-example}
\begin{split}
        \ca G \lo 1 = \ali \{  {g(y,t)} = y \hi t: t \in \natural\},  \hspace{1.14in}
        \ca G \lo 2 =  \{  {g(y,t)} = I ( y \le t): t \in \real \},   \\
        \ca G \lo 3 = \ali \{ {g(y,t)} =  {\sin(ty), \cos(ty)}: t \in \real \},  \hspace{.3in}
        \ca G \lo 4 =  \{  {g(y,t)} = \kappa (t,y): t \in \real\}, 
\end{split}
\end{align}
where $\ca G \lo 1$ defines the {general moment functions}; $\ca G \lo 2$ defines the cumulative distribution function; $\ca G \lo 3$ defines the characteristic function; and $\ca G \lo 4$, with a universal kernel $\ka $ such as the Gaussian radial basis function, defines an injective embedding of a probability into a reproducing kernel Hilbert space.   The above examples were used in \cite{yinli2011} as ensembles that characterize the central subspace. The idea of using a family of functions to characterize the central subspace for sufficient dimension reduction can be traced back to \cite{zhu2006fourier}, where a Fourier transform is used to recover the central subspace from a collection of central mean subspaces.    \cite{li2008projective} uses a family of linear transformations to perform sufficient dimension reduction for multivariate responses.  See also \cite{fertl2022ensemble} and \cite{zeng2010integral}. 

With the characteristic family (which we call ensemble) $\ca G$, we have the following equivalence: 
\begin{align*}
    Y \indep \X | \f (\X) \ \Leftrightarrow \ E[ g (Y, t) | X ] = E [ g(Y, t) | \f (\X)  ] \quad \mbox{for all  {$t \in I$}}. 
\end{align*}
Note that the right-hand side can be rewritten as $(h \lo t \circ \f ) (\X)$ for some $h \lo t: \real \hi d \to \real$. So, at the population level, we can find $\f (\X)$ by the minimizing the objective function
\begin{align}\label{eq:belted ensemble at pop level}
  \int \lo I   E \{  [ g (\Y, t) -( h \lo t \circ \f ) (\X) ] \hi 2 \} d \mu (t)
\end{align}
over a sufficiently rich family of $h \lo t \circ \f$,  {where $\mu$ is a finite measure on $I$}. To implement this optimization through the neural network family is the basic idea underlying our proposal.

\subsection{Neural network}

We first give a definition of the neural network. 
\begin{definition}\label{def:nn-func}
    A function ${\f\nn}: \real \hi p \to \real \hi d$ is called a   neural network with structural parameters $(p, l \lo 0,\k,d)$, where $\k = ( k \lo 1, \ldots, k \lo {l \lo 0})$, if it has the form
    \begin{align*}
        {\f\nn} = \A \hi {(l \lo 0+1)} \circ \sigma \circ \A \hi {(l \lo 0)} \ldots \circ \sigma \circ \A \hi {(2)} \circ \sigma \circ \A \hi {(1)}
    \end{align*}
    where, for each $l = 1, \ldots, l \lo 0+1$, $\A \hi {(l)}: \real \hi {k \lo {l - 1}}\to \real \hi {k \lo l}$ (with the convention  $k \lo 0 = p$ and $k \lo {l \lo 0+1} = d$)  is the affine function 
    \begin{align*}
        \A \hi {(l)} (\x) = \W \hi {(l)} \x + \b \hi {(l)},\quad \W \hi {(l)} \in \real \hi {k \lo {l} \times k \lo {l-1}}, \quad \b \hi {(l)} \in \real \hi {k \lo l},
    \end{align*}
and   $\sigma$ is an activation function. If,  furthermore,  $k \lo 1 = \ldots = k \lo {l \lo 0} = r \lo 0$, then we call $\f$   a   neural network with structural parameters $(p, l \lo 0,r \lo 0,d)$.
\end{definition}

In the above, for each $l=1,\ldots,l \lo 0$, the function $\sigma \circ \A \hi {(l)}$ is called the $l$-th layer of the neural network, and the last affine transformation $\A \hi {(l \lo 0+1)}$ is called the fully-connected layer. The structural parameters $p$ and $d$ are called the input and output dimensions; $l \lo 0$ is called the depth  or the number of layers of the neural network; $\k$ is called the width vector, with each $k \lo l$ called the width   of the $l$-th layer.  Furthermore, for $l = 1, \ldots, l \lo 0+1$, $\W \hi {(l)}$ and $\b \hi {(l)}$ are called the weight matrix and bias vector in the $l$-th layer, respectively, and the elements of them are called the weights and biases. In the special case where  $k \lo 1 = \cdots = k \lo {l \lo 0}$,   $r \lo 0$  is called the width of the neural network. We next define the function class of neural networks.

\begin{definition}\label{def:nn-func-class} The collection of all  neural networks with structural parameters $(p, l \lo 0, \k, d)$ and with weights and biases belonging to $[-B, B]$ for some $B > 0$ is called the neural network function class  
$\ca F\nn (p,   l \lo 0, \k, d, B)$. In the special case where $k \lo 1 = \cdots = k \lo {l \lo 0}$, we denote this class by  $\ca F\nn (p, l \lo 0, r \lo 0, d , B)$.
\end{definition}

\subsection{The belt and  ensemble of the neural network}\label{sec:belt-ensemble}

The function class we use to estimate the functions $h \lo t \circ \f$ is the composition of two neural networks, where all the hidden layers and the last layer are allowed to go to infinity with the sample size $n$ except one layer, which we call the ``belt''. The output of our neural network is usually of high dimension that allows to go to infinity with $n$, so we could capture all the features of the conditional distribution in the limit. The following definition gives an explicit description of this structure.

\begin{definition}\label{def:benn-func} Let 
 ${\f\nn}$ and ${\h\nn}$ be two neural networks with structural parameters $(p, l \lo 1, \k \lo 1, d)$ and $(d, l \lo 2, \k \lo 2, m)$, respectively,
where $\k \lo 1 = ( k \lo {11}, \ldots, k \lo {1 l \lo 1})$ and $\k \lo 2 = ( k \lo {21}, \ldots, k \lo {2 l \lo 2})$. We call the function 
    \begin{align*}
        {\h\nn} \circ {\f\nn},
    \end{align*}
a belted and ensembled  neural network  {(BENN)}. The class of all such functions,  {with all weights and biases belonging to $[-B,B]$ for some $B>0$,} is called BENN-class, and will be written as $\ca F \benn (p, l \lo 1, \k \lo 1, d, l \lo 2, \k \lo 2, m ,B)$. In the special case where the vector  $\k \lo 1$ (or $\k \lo 2)$ contains the same integer $r \lo 1$ (or $r \lo 2$),  we write the BENN-class as $\ca F \benn (p, l \lo 1, r \lo 2, d, l \lo 2, r \lo 2, m ,B)$. 
\end{definition}

Without a proper context the function ${\h\nn} \circ {\f\nn}$ is nothing but the composition of two neural networks. Nevertheless, the concepts of belt and ensemble will bring useful insights into how  our method works for sufficient dimension reduction.

The width of the belt, $d$, is the number of sufficient predictors we need to describe the regression relation. In practice, we usually choose  $d$ to be a small and   fixed (i.e. not increasing with $n$) integer, while $m$ can be a large integer. The middle layer with $d$ neurons is called  the  ``belt'', as it is usually much narrower than the layers before or after it, which constitutes the ``belly''. The sub-neural network  ${\f\nn}: \real \hi p \to \real \hi d$ functions as a  dimension reducer that brings down the original dimension of $p$ of $\X$ to a much lower number $d$, which is the dimension of the sufficient predictor. 
The sub-neural network ${\h\nn}: \real \hi d \to \real \hi m$ performs the fitting of the family of the transformations of $Y$ (or the ensemble). It attempts to capture all the information of the condition distribution of $Y|\X$. For this purpose $m$ is usually choose to be a large integer and in theory it increases with the sample size $n$. 
The next corollary shows that the $\ca F \benn$ family is a special case of the $\ca F \nn$-family. Its proof is straightforward and is omitted.

\begin{corollary} Under Definitions \ref{def:nn-func-class} and \ref{def:benn-func}, we have 
\begin{align*}
\ca F \benn ( p, l \lo 1, \k \lo 1, d, l \lo 2, \k \lo 2, m, B)= \ca F \nn ( p, l \lo 1 + l \lo 2 + 1, (\k \lo 1, d, \k \lo 2), m, B). 
\end{align*}
\end{corollary}

Our use of a narrow ``belt'' --- that is, $d$ is a small integer that does not depend on $n$ --- is consistent with statistical regression relations that are most commonly encountered in practice. Indeed, in  most regression settings,  the  predictor vector $\X$, appears only in a few features of the conditional distribution of $Y |\X$. {Specifically, two commonly used regression examples are given in Section \ref{sec:regression-example}.}  To take advantage of the  low-dimensional structure through which $\X$ enters into the conditional distribution is the main point of nonlinear sufficient dimension reduction.

The output of a BENN is fed into least-squares procedure  that fits a finite collection of transformations of $Y$. In other words, the function $\f$ in (\ref{eq:belted ensemble at pop level}) is replaced by the neural network ${\f\nn}$; the functions $\{h \lo t: t \in I \}$ is replaced by the neural network ${\h\nn}$, which has an $m$-dimensional output. Each of the components of ${\h\nn}$, say $({\h\nn}) \lo \ell$,  represent  a function that estimates  $h \lo {t \lo \ell}$, where $t \lo 1, \ldots, t \lo m$ are points in $I$. The set of $m$ functions 
\begin{align*}
  \{ ({\h\nn}) \lo \ell \circ {\f\nn}: \ell = 1, \ldots, m \}  
\end{align*} 
is the belted ensemble of neural networks that will be used as the  proxy of $h \lo t \circ \f$ in (\ref{eq:belted ensemble at pop level}). The integral $\int \lo I \cdots d \mu (t)$ will be replaced by numerical integral or sum, depending whether  $\mu$ is the Lebesgue measure or counting measure. The ensemble size $m$ is allowed to go to infinity with $n$ so that the numerical integral converges to the true integral as the sample size tends to infinity. {In some special cases, a fixed or small number $m$ would be sufficient for a consistent estimator of the central class, which is analogous to the number of slices in the sliced inverse regression method. We give some detailed discussions in Section \ref{sec:small-m}.}

\begin{figure}[!h]
    \centering
\begin{tikzpicture}[x=1.5cm, y=1.5cm, >=stealth]

\foreach \m/\l [count=\y] in {1,...,4}
  \node [circle,fill=blue!20,minimum size=1cm] (input-\m) at (0,-\y-1.5) {$x_\m$};

\foreach \m [count=\y] in {1,...,5}
  \node [circle,fill=green!20,minimum size=1cm] (hidden1-\m) at (1.5,-\y-1) {$d_{1\m}$};

\foreach \m [count=\y] in {1,...,5}
  \node [circle,fill=green!20,minimum size=1cm] (hidden2-\m) at (3,-\y-1) {$d_{2\m}$};

\foreach \m [count=\y] in {1,...,2}
  \node [circle,fill=orange!20,minimum size=1cm] (belt-\m) at (4.5,-\y-2.5) {$b_{\m}$};

\foreach \m [count=\y] in {1,...,6}
  \node [circle,fill=cyan!20,minimum size=1cm] (hidden4-\m) at (6,-\y-0.5) {$e_{1\m}$};

\foreach \m [count=\y] in {1,...,6}
  \node [circle,fill=cyan!20,minimum size=1cm] (hidden5-\m) at (7.5,-\y-0.5) {$e_{2\m}$};

\foreach \m [count=\y] in {1,...,5}
  \node [circle,fill=red!20,minimum size=1cm] (output-\m) at (9,-\y-1) {\small $g_{\m}(y)$};

\foreach \i in {1,...,4}
  \foreach \j in {1,...,5}
    \draw[->] (input-\i) -- (hidden1-\j);

\foreach \i in {1,...,5}
  \foreach \j in {1,...,5}
    \draw[->] (hidden1-\i) -- (hidden2-\j);

\foreach \i in {1,...,5}
  \foreach \j in {1,...,2}
    \draw[->] (hidden2-\i) -- (belt-\j);

\foreach \i in {1,...,2}
  \foreach \j in {1,...,6}
    \draw[->] (belt-\i) -- (hidden4-\j);

\foreach \i in {1,...,6}
  \foreach \j in {1,...,6}
    \draw[->] (hidden4-\i) -- (hidden5-\j);

\foreach \i in {1,...,6}
  \foreach \j in {1,...,5}
    \draw[->] (hidden5-\i) -- (output-\j);

\end{tikzpicture}
    \caption{A schematic plot for BENN with parameters $(p,l \lo 1, r \lo 1, d, l \lo 2, r \lo 2 , m) = (4,2,5,2,2,6,5)$, {where  $x \lo 1, \ldots, x \lo 4$ are the inputs,  $d \lo {\ell r}$'s are the neurons for dimension reduction, $b \lo 1, b \lo 2$ are neurons for the sufficient predictor,  $e \lo {\ell r}$'s are the   neurons for ensemble regression, and $g \lo 1(y), \ldots, g \lo t(y)$ are the ensemble of  transformations of $y$. }}
    \label{fig:benn}
\end{figure}
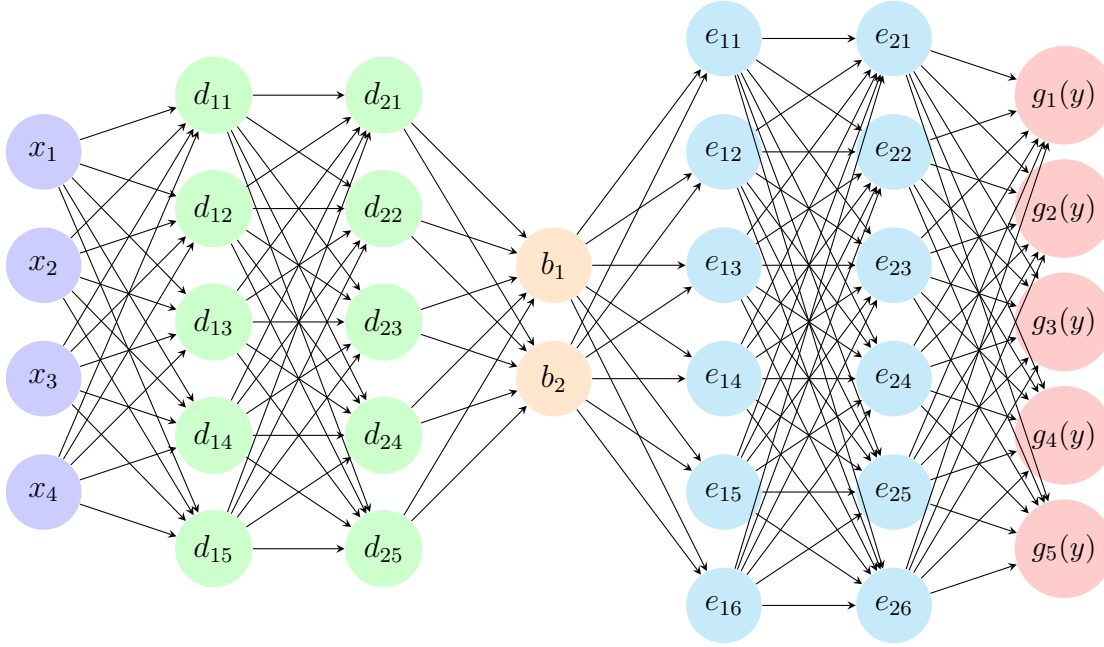

Figure \ref{fig:benn} below is a schematic plot that shows the structure of the BENN.  

\section{Varieties and  precursors}\label{sec:varieties and precursors}

\subsection{Important special cases of BENN} \label{sec:special cases}
Our proposed framework of  BENN-SDR unifies many settings in the current literature on sufficient dimension reduction, including: 
\begin{enumerate}
    \item linear and nonlinear sufficient dimension reduction,
    \vspace{-.105in}
    \item central subspace and central mean subspace for linear SDR, 
    \vspace{-.105in}
    \item central $\sigma$-field and central mean $\sigma$-field for nonlinear SDR, 
    \vspace{-.105in}
    \item  continuous and categorical responses. 
\end{enumerate} 
The great  flexibility is achieved through the choice of the ensemble, and the choice of the position of the belt.

The case shown in Figure \ref{fig:benn} corresponds to nonlinear SDR, where the sufficient predictor is the nonlinear function ${\f\nn} (\X)$, or the $\sigma$-field generated thereof. Nonlinear SDR was introduced by  \cite{lee2013}, where the nonlinear sufficient predictor was modeled by functions in the reproducing kernel Hilbert space (RKHS). Here, we model the sufficient predictor by functions in BENN. This, among other benefits,  substantially reduces the computation time, as BENN does not require inversion of large matrices, as the kernel-based method does.

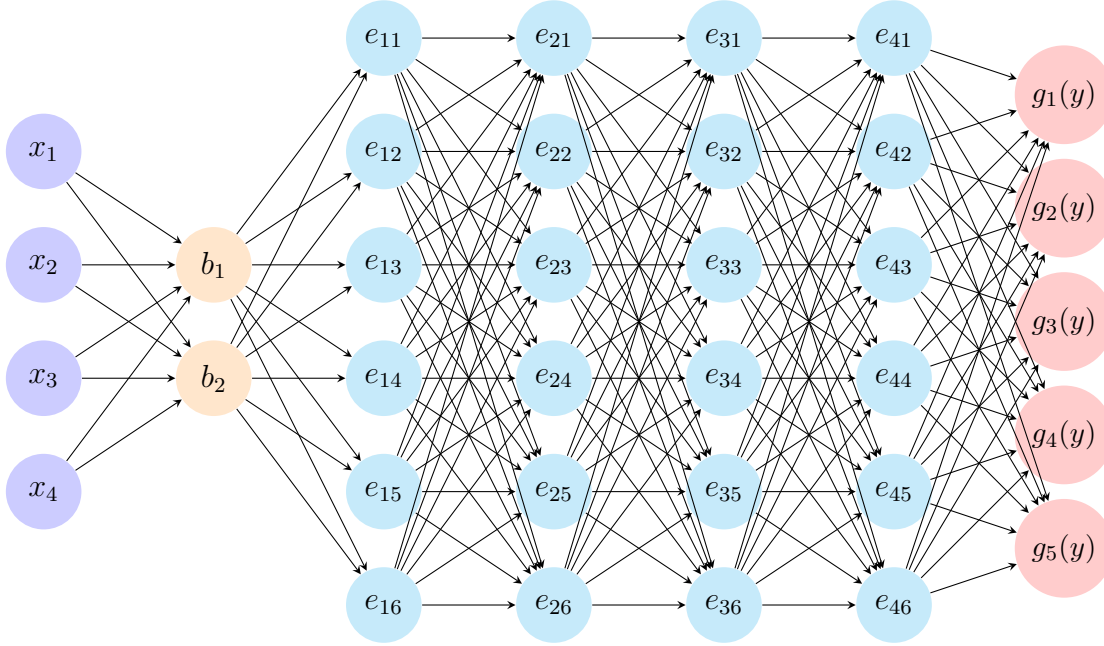
\begin{figure}[!h]
    \centering
\begin{tikzpicture}[x=1.5cm, y=1.5cm, >=stealth]

\foreach \m/\l [count=\y] in {1,...,4}
  \node [circle,fill=blue!20,minimum size=1cm] (input-\m) at (0,-\y-1.5) {$x_\m$};

\foreach \m [count=\y] in {1,...,2}
  \node [circle,fill=orange!20,minimum size=1cm] (belt-\m) at (1.5,-\y-2.5) {$b_{\m}$};

\foreach \m [count=\y] in {1,...,6}
  \node [circle,fill=cyan!20,minimum size=1cm] (hidden2-\m) at (3,-\y-0.5) {$e_{1\m}$};

\foreach \m [count=\y] in {1,...,6}
  \node [circle,fill=cyan!20,minimum size=1cm] (hidden3-\m) at (4.5,-\y-0.5) {$e_{2\m}$};

\foreach \m [count=\y] in {1,...,6}
  \node [circle,fill=cyan!20,minimum size=1cm] (hidden4-\m) at (6,-\y-0.5) {$e_{3\m}$};

\foreach \m [count=\y] in {1,...,6}
  \node [circle,fill=cyan!20,minimum size=1cm] (hidden5-\m) at (7.5,-\y-0.5) {$e_{4\m}$};

\foreach \m [count=\y] in {1,...,5}
  \node [circle,fill=red!20,minimum size=1cm] (output-\m) at (9,-\y-1) {\small $g_{\m}(y)$};

\foreach \i in {1,...,4}
  \foreach \j in {1,...,2}
    \draw[->] (input-\i) -- (belt-\j);

\foreach \i in {1,...,2}
  \foreach \j in {1,...,6}
    \draw[->] (belt-\i) -- (hidden2-\j);

\foreach \i in {1,...,6}
  \foreach \j in {1,...,6}
    \draw[->] (hidden2-\i) -- (hidden3-\j);

\foreach \i in {1,...,6}
  \foreach \j in {1,...,6}
    \draw[->] (hidden3-\i) -- (hidden4-\j);

\foreach \i in {1,...,6}
  \foreach \j in {1,...,6}
    \draw[->] (hidden4-\i) -- (hidden5-\j);

\foreach \i in {1,...,6}
  \foreach \j in {1,...,5}
    \draw[->] (hidden5-\i) -- (output-\j);

\end{tikzpicture}
    \caption{A graph of BENN for central subspace in linear SDR, with all notations same as in Figure \ref{fig:benn}.}
    \label{fig:benn-linear-cs}
\end{figure}

Another special case is the linear SDR, which can be achieved by placing the belt in the first layer of the BENN, right after the input, {without imposing an activation function.  
Thus, the belt layer includes nothing but the $d$-dimensional linear combinations of $\X$.} In this case, ${\f\nn}: \real \hi p \to \real \hi d$ is simply the linear function $\A \hii 1 \X$, and ${\h\nn} \circ {\f\nn} (\X) = {\h\nn} ( \A \hii 1 \X)$.  This means the BENN is targeting the problem 
\begin{align}\label{eq:ensemble-linear}
    E[g(Y,t)|\X] = E[g(Y,t)|\B \trans \X]
\end{align} for all $t \in I$,   which is equivalent to the problem $Y \indep \X | \B \trans \X$, the solution to which is exactly  the central subspace $\ca S \lo {Y|\X}$ in linear SDR setting. This scenario is depicted in Figure \ref{fig:benn-linear-cs}.

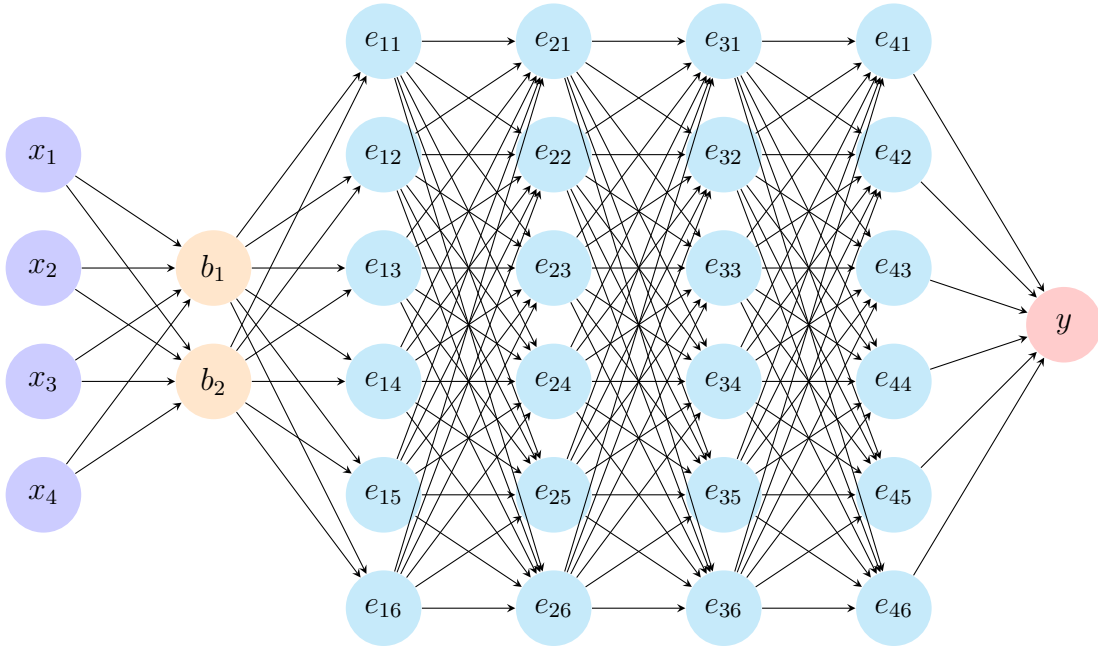
\begin{figure}[!h]
    \centering
\begin{tikzpicture}[x=1.5cm, y=1.5cm, >=stealth]

\foreach \m/\l [count=\y] in {1,...,4}
  \node [circle,fill=blue!20,minimum size=1cm] (input-\m) at (0,-\y-1.5) {$x_\m$};

\foreach \m [count=\y] in {1,...,2}
  \node [circle,fill=orange!20,minimum size=1cm] (belt-\m) at (1.5,-\y-2.5) {$b_{\m}$};

\foreach \m [count=\y] in {1,...,6}
  \node [circle,fill=cyan!20,minimum size=1cm] (hidden2-\m) at (3,-\y-0.5) {$e_{1\m}$};

\foreach \m [count=\y] in {1,...,6}
  \node [circle,fill=cyan!20,minimum size=1cm] (hidden3-\m) at (4.5,-\y-0.5) {$e_{2\m}$};

\foreach \m [count=\y] in {1,...,6}
  \node [circle,fill=cyan!20,minimum size=1cm] (hidden4-\m) at (6,-\y-0.5) {$e_{3\m}$};

\foreach \m [count=\y] in {1,...,6}
  \node [circle,fill=cyan!20,minimum size=1cm] (hidden5-\m) at (7.5,-\y-0.5) {$e_{4\m}$};

\foreach \m [count=\y] in {1}
  \node [circle,fill=red!20,minimum size=1cm] (output-\m) at (9,-\y-3) {$y$};

\foreach \i in {1,...,4}
  \foreach \j in {1,...,2}
    \draw[->] (input-\i) -- (belt-\j);

\foreach \i in {1,...,2}
  \foreach \j in {1,...,6}
    \draw[->] (belt-\i) -- (hidden2-\j);

\foreach \i in {1,...,6}
  \foreach \j in {1,...,6}
    \draw[->] (hidden2-\i) -- (hidden3-\j);

\foreach \i in {1,...,6}
  \foreach \j in {1,...,6}
    \draw[->] (hidden3-\i) -- (hidden4-\j);

\foreach \i in {1,...,6}
  \foreach \j in {1,...,6}
    \draw[->] (hidden4-\i) -- (hidden5-\j);

\foreach \i in {1,...,6}
  \foreach \j in {1}
    \draw[->] (hidden5-\i) -- (output-\j);

\end{tikzpicture}
    \caption{A graph of BENN for central mean subspace in linear SDR, with all notations same as in Figure \ref{fig:benn}.}
    \label{fig:benn-linear-cms}
\end{figure}

The third special case is linear SDR where the target of estimation is the conditional mean $E(Y|\X)$ rather than the conditional distribution of $Y$ given $\X$. This is a framework proposed in \cite{Cook2002}: we want to find the smallest subspace, say $\ca S \lo {E(Y|\X)}$, among all the subspaces  spanned by matrices $\B$ that satisfy dimension reduction relation  $E(Y|\X) = E(Y|\B \trans \X)$. The subspace $\ca S \lo {E(Y|\X)}$ is called the central mean subspace. In order to estimate $\ca S \lo {E(Y|\X)}$ by BENN, we simply replace $g(Y,t)$ in (\ref{eq:belted ensemble at pop level}) by $Y$  and choose $m=1$ in ${h\nn}$, so that the function ${h\nn} \circ {\f\nn}: \real \hi p \to \real$ in the minimization of (\ref{eq:belted ensemble at pop level}) pursues the 1-dimensional conditional mean $E(Y|\B \trans \X)$. This  neural network estimator for linear SDR was proposed earlier by \cite{kapla2022fusing}.  This scenario is depicted schematically by   Figure \ref{fig:benn-linear-cms}.

Further developing along these lines,  \cite{yin-cook-2002} introduced the notion of $k$-th central moment space, where we aim to estimate  the smallest subspace, say $\ca S \lo {E(Y \hi k | \X)}$, among all the subspaces spanned by the matrices $\B$ that satisfy $E(Y \hi j|\X) = E(Y \hi j|\B \trans \X)$ for $j=1,\ldots,k$. This subspace is called the $k$-th central moment subspace. To estimate $\ca S \lo {E(Y \hi k |\X)}$ by BENN, we take $g(Y, \ell) = Y \hi \ell$ for {$\ell = 1,\ldots, k$} and set the dimension of the output of ${\h\nn}$ to be $k$, so that {the $\ell$-th} component of ${\h\nn} ( \A \hii 1 \X )$ can target $E(Y \hi \ell | \X  )$.  

Furthermore, when the response $\Y$ is multivariate, we can also consider the central mean $\sigma$-field that is sufficient for estimating $E(\Y|\X)$ nonlinearly. That is, given the responses $\Y \in \real \hi q$ and covariates $\X \in \real \hi p$, we seek a nonlinear function $\f:\real \hi p \to \real \hi d$ with $d<q$ such that $E(\Y|\X) = E[\Y|\f(\X)]$. In this case, we can place  $Y \lo 1, \ldots, Y \lo q$ on the output layer, and apply the neural network as in Figure \ref{fig:benn}.

Finally, consider the case where  $Y$ is a categorical random variable, which, without loss of generality, can be assumed to take values in a finite set $\{1, \ldots, K \}$. We  can perform nonlinear SDR on such a $Y$ by choosing  the ensemble to be the set of indicator functions of each category: $\{\1 \lo {\{y=1\}}, \ldots, \1 \lo {\{y=K\}} \}$ and setting the dimension of the output of the neural network ${\h\nn}$ to be $K$. In this way we can directly recover the conditional distribution of $Y|\X$, since these indicator functions {characterize the distribution when $Y$ is categorical.}

\subsection{Relation with autoencoder }

Apart from the mentioned paper by \cite{kapla2022fusing}, another precursor of our method is what is known as the autoencoder for unsupervised dimension reduction. 
\cite{kramer1991nonlinear} introduced a special  type of neural network, the auto-associative neural network,   to conduct nonlinear principal component analysis. They constructed a   neural network,  with a narrow layer in the middle,   to fit $\X$ itself. Since this is using a function of $\X$ as the predictor to fit $\X$ as the response, the authors call their method the ``autoencoder''. The low-dimension random vector in the narrow layer is used as the nonlinear principal components of $\X$. 
See also \citet{bourlard1988auto} and \citet{hinton1993autoencoders} for related developments. {Moreover, \citet{zhong2023nonlinear} generalizes the autoencoder method to nonlinear functional principal component analysis by constructing a transformed functional autoassociative neural network, where the functions are expanded under a B-spline basis.}  

The  narrow layer in the autoencoder plays the role of the belt in our estimator. The difference between our method and the autoencoder  is that we use a belted network to fit a class of functions of $Y$ rather than $\X$ itself. Thus, the autoencoder is a unsupervised dimension reduction method, whereas BENN-SDR is a supervised dimension reduction method. The common point is that both methods use a narrow layer in a neural network as the vehicle for  dimension reduction. In a sense, the relation between our proposed BENN-SDR and the autoencoder resembles the relation  between nonlinear SDR based on reproducing kernel Hilbert space (\cite{lee2013}) and the kernel principal component analysis (\cite{scholkopf1998nonlinear}).

\section{Implementation via minimization of sum of squares}\label{sec:implementation}

\subsection{Fisher consistency}

In this section we develop the sample-level implementation of the   BENN-SDR method. To motivate our numerical procedure, we first show that the minimization of the population-level criterion (\ref{eq:belted ensemble at pop level}) indeed gives the solution to the dimension reduction problem (\ref{eq:nonlinear-sdr-cs}). Let $P \lo {\X}$ be the distribution of $\X$, and $\mu$ be a finite measure on $I$ (as appeared in (\ref{eq:ensemble})). Let $L \lo 2 (P \lo {\X})$ be the collection of all measurable  functions of $\X$ square-integrable with respect to $P \lo {\X}$, and let $L \lo 2 (\mu \times P \lo {\X})$ be the collection of all measurable functions of $(t, \X)$ square-integrable  with respect to $\mu \times P \lo {\X}$. Let $[L \lo 2 (P \lo {\X})] \hi d$ be the Hilbert space consisting of the $d$-fold Cartesian product of $L \lo 2 (P \lo {\X})$ and the inner product 
\begin{align*}
    \langle \u, \v  \rangle = \sum \lo {i=1} \hi d E [u \lo i (\X) v \lo  i (\X) ]. 
\end{align*}
where $\u = (u \lo 1, \ldots u \lo d)$ and $\v = (v \lo 1, \ldots, v \lo d)$ are members of $d$-fold Cartesian product of $L \lo 2 (P \lo {\X})$.

\begin{theorem}\label{thm:population}
Suppose 
\begin{enumerate}
    \item  $\ca G = \{ g(\cdot, t): t \in I \}$
    characterizes the distribution of $Y$, and $\mu$ is a finite measure on $I$; 
    \vspace{-.08in}
    \item {$g(y,t)$ is uniformly bounded for all $(y,t) \in \Omega \lo Y \times I$, and is continuous in $t$ on $I$;}
    \vspace{-.08in}
    \item $\ca H \lo 1$ is a subset of $[L \lo 2( P \lo {\X})] \hi d$; 
     \vspace{-.08in}
    \item there is a member {$\f \lo 0$}  of $\ca H \lo 1$ such that $Y \indep \X | {\f \lo 0} (\X)$; 
     \vspace{-.08in} 
    \item $\ca H \lo 2$ is a collection of real-valued functions on $I \times \Omega \lo {\X}$ to $\real$ that are of the form $h {[\tilde \f (\X),t]} $ with $\tilde \f \in \ca H \lo 1$ and that are members of {$L \lo 2 ({P \lo {\X} \times \mu})$;} 
     \vspace{-.08in}
    \item the function $({\X, t}) \mapsto E [ g (Y, t) | {\f \lo 0}(\X)]$ is a member of $\ca H \lo 2$.
\end{enumerate} 
If {$h [\f(\X),t]$} is a minimizer of 
    \begin{align*}
           \int E \left\{  g(Y, t)   - {v [\u(\X),t]} \right\}^2 d\mu(t)  
    \end{align*}
    over $\ca H \lo 2$, then 
      {$\f$}   satisfies the SDR condition  $Y \indep \X |{\f}(\X)$.
\end{theorem}

{In general, $\f$ has the same dimension as the function that generates the central $\sigma$-field, but may generate a larger $\sigma$-field than the central $\sigma$-field. In an important special case, though,  $\f$ generates the central $\sigma$-field. A rigorous investigation of the relation between $\f$ and the central $\sigma$-field is given in in Section \ref{sec:sigma-field} in the Supplementary Material.}

\subsection{Objective function  at  sample level}\label{sec:objective}

For convenience in practice and proof, we set the support of $t$ to be a bounded interval $I = [0,\tau]$. 
{
Some examples in \eqref{eq:ensemble-example} can be restricted to bounded intervals, which are discussed in Section \ref{sec:supp-t}. 
}
We choose $\mu$ and $t \lo 1, \ldots, t \lo m$ according to  the following assumption.

\begin{assumption}\label{ass:t-mu}
    $0 = t \lo 1 < \ldots < t \lo m < t \lo {m+1} = \tau$ are equally spaced grid points in $I$, and $\mu$ is the Lebesgue measure on $I$.
\end{assumption}

{We also make the following assumption on the ensemble, which will be used for convenience of proof.}

\begin{assumption}\label{ass:bound-fy}
    The ensemble  $\{g(y,t): t \in I \}$ is uniformly bounded by $B \lo y$ for all $(y,t) \in \Omega \lo Y \times I$.
\end{assumption}

{This is a mild assumption: for example,  the ensembles $\ca G \lo 2, \ca G \lo 3, \ca G \lo 4$ in \eqref{eq:ensemble-example} all satisfy this condition, and $\ca G \lo 1$ also satisfies it for bounded $Y$ after proper scaling.}
Before describing  our estimator, we introduce the truncation functional $T \lo B$.
\begin{definition}
    For any $B>0$  and  any function $f: \real \hi p \to \real$, the truncation functional $T \lo B$ is defined as
    \begin{align*}
       ( T \lo B f ) (\x) = 
        \begin{cases}
            B, & f(\x) > B, \\
            f(\x), & -B \le f(\x) \le B,\\
            -B, & f(\x) < -B.
        \end{cases}
    \end{align*}
For a vector-valued function  $\f = (f \lo 1, \ldots, f \lo d) \trans$, we define  $T \lo B \f = (T \lo B f \lo 1, \ldots, T \lo B f \lo d) \trans$.
\end{definition}

Using this notation, and under Assumption \ref{ass:t-mu}, we propose our estimator as
\begin{align}\label{eq:ghat-star}
    {(\hat{\h} \hi *, \hat{\f}) = (T \lo {B \lo y} \hat{\h}, \hat{\f}),}
\end{align}
where, {for some $B \lo w >0$,}
\begin{align}\label{eq:ghat-vec}
\begin{split}
    {(\hat{\h},\hat{\f})}
    = \ali  {\argmin} \left\{ \sum \lo {j=1} \hi m \sum \lo {i=1} \hi {n} | g (Y \lo i, t \lo j ) - {T \lo {B \lo y} h \lo j \circ \f}  ( \X \lo i ) | \hi 2 : \right. \\
    \ali  \hspace{.8in}\left. \phantom{\sum \lo {j=1} \hi m}
    {\f \in \ca F \nn ( p, l \lo 1, r \lo 1, d), \h \in \ca F \nn (d, l \lo 2, r \lo 2, m)} \right\}.
\end{split}
\end{align}
{Note that the condition in \eqref{eq:ghat-vec} is equivalent to $\h \circ \f \in   \ca F \benn ( p, l \lo 1, r \lo 1, d, l \lo 2, r \lo 2, m, B \lo w)$ for $\f: \real \hi p \to \real \hi d$ and $\h: \real \hi d \to \real \hi m$.}
The estimated sufficient predictors for $Y \lo 1, \ldots, Y \lo n$ are the $d$-dimensional vectors $\{\hat{\f} (\X \lo i ): i = 1, \ldots, n \}$. 
{For notation convenience, we omit the subscript $\,\nn$ in the estimated functions because all of them are restricted in the neural network function space.}

The sum $\sum \lo {j=1} \hi m$ in the objective function in  (\ref{eq:ghat-vec}) is designed as an approximation of the integral $\int \lo I$ in (\ref{eq:belted ensemble at pop level}). Ideally, we would like to construct a  neural network that produces uncountably many outputs, each corresponding to a member of the ensemble $g(Y,t)$.   Since  this is impossible, we replace the integral with a sum and make $m$ goes to infinity with the sample size. To make this point rigorous, let $\tilde g  (Y, \cdot)$ and $\tilde f (\X, \cdot)$ be the piecewise constant functions
\begin{align*}
    \tilde g(Y, t) = \sum \lo {j=1} \hi m g (Y, t \lo j) I ( t \lo j \le t < t \lo {j+1}), \quad 
    {\tilde h[\f(\X),t] = \sum \lo {j=1} \hi m   T \lo {B \lo y} h \lo j \circ \f (\X) I ( t \lo j \le t < t \lo {j+1}). }
\end{align*}
Then the double sum in (\ref{eq:ghat-vec}), after rescaled by $n \inv m \inv \tau$,  can be rewritten as 
\begin{align}\label{eq:ghat-t-def}
\int \lo I E \lo n {\left( \{  \tilde g (Y, t) - \tilde h [\f(\X), t] \} \right)\hi 2 } d \mu (t), 
\end{align}
where, for a function $u(\X, Y)$,  $E \lo n u(\X, Y)$ is the sample average of $\{u (\X \lo i, Y \lo i): i = 1, \ldots, n \}$, and $\mu$ is the Lebesgue measure. We see that the above objective function closely resemble its population-level counterpart (\ref{eq:belted ensemble at pop level}).

\subsection{Implementation algorithm}\label{subsec:implementation}

We use PyTorch to implement our method,  following the framework outlined in Section 10.9 of \cite{james2023introduction}.  Recall that our BENN is of the form  {$\h\nn \circ \f\nn$}.
The dimension-reduction neural network {$\f\nn$}, which  maps the $p$-dimensional vector $\x$ to a $d$-dimensional vector  $\z = {\f\nn}(\x)$,   has  $l \lo 1$ layers, with its  $i$-th layer having  $k \lo {1i}$ neurons. The ensemble neural network ${\h\nn}$, which maps the $d$-dimensional vector $\z$ to an $m$-dimensional vector $\tilde{\y}$ as a prediction of the transformed response $\g(y) \equiv [g(y,t \lo 1), \ldots, g(y,t \lo m)] \trans$,  has $l \lo 2$ layers, with its  $i$-th layer having  $k \lo {2i}$ neurons. 
Let {$(\hat \h, \hat \f)$} be the solution to the optimization problem (\ref{eq:ghat-vec}). We set the  output of our procedure to contain both  the $d$-dimensional sufficient predictors $\hat \z \lo i  = {\hat \f}(\x \lo i )$, $i = 1, \ldots, n$,   and the $m$-dimensional predicted values of the ensemble $\{ {\hat \h \circ \hat \f} (\x \lo i) : i = 1, \ldots, n \}$. Algorithm \ref{alg:dr} below gives the  forward pass algorithm for calculating  ${\hat \h \circ \hat \f}$.

\begin{algorithm}[h]
{\small
\caption{Forward pass for ${\hat \h \circ \hat \f}$ in BENN}
\label{alg:dr}
\begin{algorithmic}[1]
\Require Input data $\x \in \real \hi p$; 
\Require Structure for {$\hat \f$}: number of layers $l \lo 1$, width vector $\k \lo 1= (k \lo {11}, \ldots, k \lo {1 l \lo 1})$, weights $\{\W \hi {(1i)} \} \lo {i=1} \hi {l \lo 1+1}$, biases $\{\b \hi {(1i)} \} \lo {i=1} \hi {l \lo 1 +1}$, where $k \lo {10}=p$ and $k \lo {1(l \lo 1+1)} =d$; 
\Require Structure for {$\hat \h$}: number of layers $l \lo 2$, width vector $\k \lo 2= (k \lo {21}, \ldots, k \lo {2 l \lo 2})$, weights $\{\W \hi {(2i)} \} \lo {i=1} \hi {l \lo 2+1}$, biases $\{\b \hi {(2i)} \} \lo {i=1} \hi {l \lo 2 +1}$, where $k \lo {20}=d$ and $k \lo {2(l \lo 2+1)} =m$;
\Ensure Output $\z = {\hat{\f}} (\x)$ and
$\tilde{\y} = {\hat{\h}}(\z)$  of the neural network;

\State $\mathbf{a} \hi {(10)} \gets \mathbf{x}$ 

\For{$i = 1$ to $l \lo 1$}
       \State $\mathbf{a} \hi {(1i)} \gets \sigma (\mathbf{W} \hi {(1i)} \mathbf{a} \hi {(1,i-1)} + \mathbf{b} \hi {(1i)})$ 
\EndFor

\State $\z \gets \mathbf{W} \hi {(1,l \lo 1 +1)} \mathbf{a} \hi {(1l \lo 1)} + \mathbf{b} \hi {(1,l \lo 1 +1)}$
\Comment{$\z$ is the dimension reduction result ${\hat \f} (\x)$}

\State $\mathbf{a} \hi {(20)} \gets \mathbf{z}$ 

\For{$i = 1$ to $l \lo 2$}
        \State $\mathbf{a} \hi {(2i)} \gets \sigma (\mathbf{W} \hi {(2i)} \mathbf{a} \hi {(2,i-1)} + \mathbf{b} \hi {(2i)})$ 
\EndFor

\State $\tilde{\y} \gets \mathbf{W} \hi {(2,l \lo 1 +1)} \mathbf{a} \hi {(2l \lo 1)} + \mathbf{b} \hi {(2,l \lo 1 +1)}$
\Comment{$\tilde{\y}$ is the ensemble regression result ${\hat \h} (\z)$}

\Return $\z$, $\tilde{\y}$ 
\end{algorithmic}
}
\end{algorithm}

For the loss function, we directly apply the $L \lo 2$-loss between the output vector $\tilde{\Y}$ and the {transformed response $\g(y)$,  and we perform back propagation to minimize the loss function}. The loss function is implemented by \texttt{MSELoss()} in PyTorch.

\section{Convergence rate of BENN}\label{sec:rate}

\subsection{Notations and assumptions}

Let
\begin{align}\label{eq:def-gtx}
    s (\x, t) = E[g (Y, t) | \X = \x].
\end{align}
Note that, under the dimension reduction assumption $Y \indep \X | {\f}  (\X)$, the above function can be rewritten as 
\begin{align*}
  s ( \x, t) = h [{\f} (\x),t] = h (\cdot,t) \circ {\f}  (\x). 
\end{align*}
For a vector $\t = (t \lo 1, \ldots, t \lo m ) \trans$, any $\u \in \real \hi d$, $\x \in \real \hi p$ and $y \in \real$  let 
\begin{align}\label{eq:h s g}
\begin{split}
\h (\u) =\ali  [ h (\u, t \lo 1), \ldots, h (\u, t \lo m) ] \trans, \\ 
\s   (\x ) =\ali  [s  ( \x, t \lo 1), \ldots, s (\x, t \lo m)] \trans, \\
    \g (y ) = \ali  [g (y, t \lo 1), \ldots, g (y, t \lo m)] \trans.
\end{split}
\end{align}
Then, we can write
\begin{align}\label{eq:gt-form-composition}
   \s   = \h  \circ {\f}
\end{align}
where ${\f}$ is a function from $\real \hi p$ to $\real \hi d$, and $\h$ is a function from $\real \hi d$ to $\real \hi m$.

We make the following additional assumptions.

\begin{assumption}\label{ass:support-x}
    The support   $\Omega \lo \X$ of $\X$  is bounded by the rectangle  $[-B \lo x, B \lo x] \hi p$.
\end{assumption}

\begin{assumption}\label{ass:lip-t}
    The conditional expectation function $s(\x,t)$, as defined in \eqref{eq:def-gtx}, is uniformly Lipschitz continuous in $t$;  that is,  there exists a constant $L \lo s >0$ such that,  for all $\x \in \Omega \lo \X$ and all $t, t' \in I$, 
\begin{align*}
    | s  (\x, t ) - s (\x, t' ) | \le L \lo s |t - t'|.
\end{align*}
\end{assumption}

We will use $h \lo 1, \ldots, h \lo m $ to denote the   components of $\h$, and use {$f \lo 1, \ldots, f \lo d$} to denote the components of  ${\f}$. 

\begin{assumption}\label{ass:ge-lip}
For each $m = 1, 2, \ldots$, the functions  {$\{h \lo j : j = 1, \ldots, m \} $ }  are  Lipschitz continuous with a universal  constant $L \lo h$; that is, for all $\x, \x' \in \real \hi d$ and all $j = 1, \ldots, m$ and all $m = 1, 2, \ldots $,
\begin{align*}
    | h \lo j (\x)  - h \lo j (\x') | \le L \lo h \| \x - \x' \|.
\end{align*}
\end{assumption}

{Assumptions \ref{ass:lip-t} and \ref{ass:ge-lip} are reasonably mild to impose on ensembles and models. Illustrative examples are provided for them in Section \ref{sec:example-ass}.}

\begin{assumption}\label{ass:gd-lip}
    The functions  {$\{ f  \lo k : k = 1, \ldots, d \}$} are  Lipschitz continuous with a universal  constant $L \lo f$; that is, for all  ${k}=1,\ldots,d$ and  for all $\x, \x' \in \real \hi p$, 
\begin{align*}
    |{f \lo k} (\x) - {f  \lo k} (\x')| \le L \lo f \| \x - \x' \|.
\end{align*}
\end{assumption}

{In addition, when investigating the convergence rate, we set the activation functions in all neural networks to be the ReLU function; that is, $\sigma(x) = x \hi + \equiv \max\{0,x\}$.}

\subsection{Convergence rate}

Since the goal of nonlinear SDR is the extract the sufficient predictor $\hat \f (\X)$, ideally, to assess the accuracy of nonlinear SDR,  one should target the distance between $\hat \f$ and $\f$ in (\ref{eq:nonlinear-sdr-cs}). However, note that the identifiable parameter in (\ref{eq:nonlinear-sdr-cs}) is the $\sigma$-field generated by $\f (\X)$: if $\f \lo 1 (\X)$ is a one-to-one transformation of $\f (\X)$ then (\ref{eq:nonlinear-sdr-cs}) continues to hold for $\f \lo 1$. Thus it is meaningless to compare the functions $\f$ and $\f \lo 1$ themselves. Instead, we should use  a numerical measurement of the difference of two $\sigma$-fields, if this is possible.

Consider two statistics $\T \lo 1$ and $\T \lo 2 $ that are measurable with respect to $\sigma (\X)$ and are of the same dimension. Moreover, suppose that $\ca A$ is the smallest $\sigma$-field such that $\Y \indep \X | \ca A$ and that $\ca A$ is generated by $\T \lo 1$. Then,   $\sigma ( \T \lo 2 ) = \ca A$ if and only if 
\begin{align}\label{eq:difference of sigma fields}
E [ g(Y,t) | \T \lo 2] = E [ g(Y,t) | \X].   
\end{align} 
So it is reasonable to measure the distance between the $\sigma$-fields generated by $\T \lo 1$ and $\T \lo 2$ by the conditional expectations $E [ g(Y,t) | \X]$ and $E [ g(Y,t) | \T \lo 2]$. This motivates us to use the following criterion  to measure the accuracy of the estimated central $\sigma$-field: 
\begin{align}\label{eq:sqloss-def}
    \ca L = E \left[ \iint \left| E[ g (Y, t) | \X = \x] - {\hat{h} \hi * [\hat{\f}(\x), t]} \right| \hi 2 d \mu (t) d P \lo \X (\x) \right]
\end{align}
where 
\begin{align}\label{eq:piecewise-constant-hat-def}
    {\hat{h} \hi * [\hat{\f}(\x), t]} = \sum \lo {j=1} \hi m {\hat{h} \hi * \lo j \circ \f }(\x) I (t \lo j \le t < t \lo {j+1} ),
\end{align}
and $\mu$ is the Lebesgue measure on $I$.  In \eqref{eq:sqloss-def}, the outer expectation is taken with respect to the observed data $\{(\X \lo 1, Y \lo 1), \ldots, (\X \lo n, Y \lo n) \}$, which appears in the expression only through the symbol ``hat'' in {$\hat \h \hi *$ and $\hat \f$}. From \eqref{eq:ghat-t-def} we see that {$\hat{h} \hi * [\hat{\f}(\x), t]$} is the truncated least squares estimate of $g(Y,t)$; so it can be viewed as the best estimate of the conditional mean $E[g(Y,t)|\X]$. The criterion (\ref{eq:sqloss-def}) is consistent with our consideration that leads to (\ref{eq:difference of sigma fields}): the central $\sigma$-field plays the role of $\ca A$; the family $\{ {h[\f(\X), t]}: t \in I \}$ plays the role of $\T \lo 1$, and the estimated family $\{ {\hat{h} \hi * [\hat{\f}(\X), t]}: t \in I \}$ plays the role of $\T \lo 2$. 

The convergence rate of $\ca L$ is given by the next theorem.  {This is the deep learning regression using $g(Y,t)$ as the response and $\X$ as predictors, and the convergence rate $\ca L$ is the mean squared error of this nonparametric regression.} In the following, we use $a \lo n \lesssim b \lo n$ to represent $a \lo n \le c b \lo n$ where $c$ is a constant that does not depend on $n$. If $a \lo n \lesssim b \lo n$ and $b \lo n \lesssim a \lo n$, then we write $a \lo n \asymp b \lo n$.

\begin{theorem}\label{thm:conv-rate}
Suppose that 
\begin{enumerate}
    \item $\t=(t \lo 1, \ldots, t \lo m) \trans$ satisfies Assumption \ref{ass:t-mu};
    \vspace{-.07in}
    \item the ensemble $\{g(y,t), t \in I \}$ satisfies Assumption \ref{ass:bound-fy} with $B \lo y \ge 1$;
    \vspace{-.07in}
    \item the support of $\X$ satisfies Assumption \ref{ass:support-x};
    \vspace{-.07in}
    \item the conditional expectation function $s(\x,t)$ defined in \eqref{eq:def-gtx} satisfies Assumption \ref{ass:lip-t};
    \vspace{-.07in}
    \item $\s$ has the form \eqref{eq:gt-form-composition}, where $\h = (h \lo 1, \ldots, h \lo m)$ and ${\f = (f \lo 1, \ldots, f \lo d)}$ satisfy Assumptions \ref{ass:ge-lip} and \ref{ass:gd-lip}, respectively.
\end{enumerate}  
Let $\ca L$ be defined by \eqref{eq:sqloss-def} and \eqref{eq:piecewise-constant-hat-def}, where {$(\hat \h \hi *, \hat \f)$} is the minimizer defined by \eqref{eq:ghat-star} and \eqref{eq:ghat-vec}. Let $L \lo 1, L \lo 2, N \lo 1, N \lo 2$ be positive integers, $B \lo w$ be a sufficiently large number such that $B \lo w \ge B \lo y$ and $B \lo w \asymp n$, and let the parameters $l \lo 1, r \lo 1, l \lo 2, r \lo 2$ in \eqref{eq:ghat-vec} be
\begin{align}
\begin{split}\label{eq:choice-l-r}
    \ali l \lo 1 = 12L \lo 1+14,  \quad r \lo 1 = \max\{4p\lfloor N \lo 1 \hi {1/p} \rfloor +3p, 12d N \lo 1+8d\}, \\
    \ali l \lo 2 = 12L \lo 2+14,  \quad r \lo 2 = \max\{4d\lfloor N \lo 2 \hi {1/d} \rfloor +3d, 12m N \lo 2+8m\}.
\end{split}
\end{align}
Then, we have
    \begin{align}\label{eq:l-bound-final}
    \begin{split}
    \ca L \lesssim \ali m \hi {-2} + n \inv (L \lo 1 N \lo 1 \hi 2 + L \lo 2 N \lo 2 \hi 2 m \hi 2) ( L \lo 1  \log N \lo 1 + L \lo 2  \log m + L \lo 2 \log N \lo 2 \\
     \ali + L \lo 1 \log n + L \lo 2 \log n    )  +  N \lo 1 \hi {-4/p} L \lo 1 \hi {-4/p} +  N \lo 2 \hi {-4/d} L \lo 2 \hi {-4/d}.
    \end{split}
\end{align}
\end{theorem}

\medskip 

While the detailed proof of the theorem is given in the Supplementary Material, we outline here the main ideas and the important references used. We use the general result 
in \cite{gyorfi2002distribution} to link  the convergence rate of nonparametric regression with the covering number of the class of nonparametric functions used, and use 
\cite{shen2024exploring} to develop the covering numbers   specific to structure of our BENN family. We use \cite{shen2020deep} to analyze  the bias caused by approximating Lipschitz functions from within a deep learning family. We follow the ideas  employed by \citet{bagirov2009estimation}, \citet{bauer2019deep} and \citet{kohler2021rate} to decompose the  mean squared regression error. 

{On deriving Theorem \ref{thm:conv-rate}, Assumption \ref{ass:support-x} is an essential condition which is commonly used, and we provide some discussions about it in Section \ref{sec:dis-support}.
We also give some discussions regarding the bounds of the weights and biases in the neural networks in Section \ref{sec:dis-bound-weight}. }

\subsection{Optimal rates of tuning parameters}\label{sec:optimal-rate}

Note that the convergence rate in Theorem \ref{thm:conv-rate} explicitly depends on the neural network parameters $L \lo 1, L \lo 2, N \lo 1, N \lo 2$ and ensemble size  $m$. This allows us to    derive the optimal choice of these tuning parameters to optimize the  convergence rate  in Theorem \ref{thm:conv-rate}.

\begin{theorem}\label{thm:optimal-rates}
    Suppose that  the conditions in Theorem \ref{thm:conv-rate} are satisfied and  that the tuning parameters  $m, L \lo 1, L \lo 2, N \lo 1, N \lo 2$ are of the form 
    \begin{align}\label{eq:tuning-para-form}
        m \asymp n \hi \alpha, \quad L \lo 1 \asymp n \hi {\beta \lo 1}, \quad L \lo 2 \asymp n \hi {\beta \lo 2}, \quad N \lo 1 \asymp n \hi {\gamma \lo 1}, \quad N \lo 2 \asymp n \hi {\gamma \lo 2}
    \end{align} 
    for some $\alpha, \beta \lo 1, \beta \lo 2, \gamma \lo 1, \gamma \lo 2 {\ge 0}$. When $d \le p-2$, the optimal rate of $\ca L$ up to a log factor in \eqref{eq:l-bound-final} is
    \begin{align}\label{eq:optimal-rate}
        \ca L \lesssim  n \hi {-\frac{2}{p+2}} \log n ,
    \end{align}
   and this rate can be achieved by the following choice of the tuning parameters 
\begin{align}\label{eq:tuning-para-choice}
        \alpha = \frac{p}{(p+2)(d+2)}, \quad {\beta \lo 1 + \gamma \lo 1 = \frac{p}{2p+4}, \quad \beta \lo 1 = \beta \lo 2, \quad \gamma \lo 1 = \gamma \lo 2  + \alpha.}
    \end{align}
\end{theorem}

Under  the optimal choice of tuning paramaters, we have $m \asymp n \hi {\frac{p}{(p+2)(d+2)}}$, which is always of a smaller order of magnitude than $n$. That means  we do not need to perform as many as $n$ transformations on $Y$. In comparison, in a kernel nonlinear SDR method, such as GSIR (\cite{lee2013}), we effectively applied $n$ transformations on $Y$:  $\ka(\cdot,Y \lo 1), \ldots, \ka(\cdot,Y \lo n)$.

 Furthermore, we show that the convergence rate of BENN is faster than that of the neural network regression  without dimension reduction. Due to space limit, we place this result in the Supplementary Material.

\section{Simulations}\label{sec:simulation}

In this section we compare our BENN method with several existing nonlinear and linear dimension reduction methods, including a kernel based method and three other neural network based methods. We will denote the regression model of $Y$ versus $\X$ by roman letters A, B, C, \ldots, and the models for the distribution of $\X$ by roman numbers $\rom{1}$, $\rom{2}$, $\rom{3}$, \ldots. To assess  accuracy, we use the distance correlation (\cite{szekely2007measuring}) for nonlinear SDR, and the distance between two projection matrices for linear SDR.  {For each model setting below, before the full-scale application of  BENN, we first conduct preliminary experiments on two extra datasets   with sample size $n=5000$  to pick a reasonable number of iterations   to avoid severe overfitting. This iteration number is then fixed throughout the rest of the simulations for this model. } Our simulation studies contain four settings: nonlinear SDR without heteroscedasticity, nonlinear SDR with heteroscedasticity, nonlinear SDR with heteroscedastity alone, linear SDR with heteroscedasticity. 
Due to space limit, we present the second setting in this section and place all the other settings   in Section \ref{sec:additional-simulation}.

For the setting of nonlinear SDR with heteroscedasticity, we consider a setting where $\X \in \real \hi p$ with $p=50$ and $d = 2$. The regression model is 
\begin{align*}
   \mathrm{D}: \quad Y = [\sin((X \lo 1 + X \lo 2)\pi/10)+X \lo 1 \hi 2] + [2 \sin \hi 2((X \lo 3+X \lo 4)\pi/10)+X \lo 3 \hi 2] \epsilon,
\end{align*}
where $\epsilon$ is a standard normal random variable independent of $\X$,  
and $\X$ is generated by 
\begin{align*}
\rom{4}: \quad     X \lo 1, \ldots, X \lo p \  \overset{\mathrm{i.i.d.} }  {\sim} \ N(0.2,0.5).
\end{align*}
The sufficient predictors for $Y$ are any one-to-one function of  
\begin{align*}
    \f(\X) = [f \lo 1 (\X),\  f \lo 2 (\X)] = [\sin((X \lo 1 + X \lo 2)\pi/10)+X \lo 1 \hi 2,\  2 \sin \hi 2((X \lo 3+X \lo 4)\pi/10)+X \lo 3 \hi 2].
\end{align*} 
In this regression model,   $E(Y|\X)$ is  a function of $f \lo 1 (\X)$, $\var(Y|\X)$ is a function of $f \lo 2 (\X)$, and the conditional distribution $P \lo {Y|\X}$ depends on $\X$ only through $E(Y|\X)$ and $\var (Y|\X)$. This  is a widely used regression hypothesis known as regression with heteroscedasticity.

We choose the training-set sample sizes to be  {$n=1000,\dots,8000$,} and the testing-set sample size to be  1000.
We compare our BENN method with the same set of alternative methods. For the BENN method, we {take $m=1,2$ combined with the ensemble family $\ca G = \{g(y)=y\}$ and $\ca G = \{ g \lo 1(y) = y, g \lo 2 (y) = y \hi 2 \}$, respectively, as well as  $m=1000$ combined with the Gaussian kernel function class as the  ensemble,  which is}
$    \ca G = \{ g(y, y \lo k)  = \exp [ -( y \lo k - y) \hi 2 / (2 \sigma \hi 2) ] , \  k=1,\ldots,m\}$, 
where $y \lo 1, \ldots, y \lo m$ are prespecified constants  generated independently  from independent uniform distribution $U(\hat{\mu} \lo Y - 2 \hat{\sigma} \lo Y, \hat{\mu} \lo Y + 2 \hat{\sigma} \lo Y)$, with $\hat{\mu} \lo Y$ and $\hat{\sigma} \lo Y$ being  the sample mean and sample standard deviation of $Y \lo 1, \ldots Y \lo n$. The bandwidth $\sigma$ in {$\ca G$} is also set to be $\hat{\sigma} \lo Y$. We choose  the structural parameters for BENN  as $(p, l \lo 1, r \lo 1, d, l \lo 2, r \lo 2, m)=(50,2,50,1,1,2000,1000)$ and use 150 epochs. For the  GMDDNet, \cite{chen2024deep} propose two procedures, one based on the successive procedure (denoted by GMDDNet-S) and the other based on the Frobenius norm (denoted by GMDDNet-F), and the GMDDNet-S shows a slightly better performance in \cite{chen2024deep}. In our simulations we use the successive procedure (GMDDNet-S) under the proposed settings. Regarding the StoNet method, we adopt a one-hidden-layer StoNet with 25 hidden units, and change the default \texttt{sigma\_list} from $(10 \hi {-3}, 10 \hi {-6})$ to $(10 \hi {-2}, 10 \hi {-4})$ to accommodate the data scale. For the GSIR, we use the default parameters in \cite{li2018sufficient}. Note that {we disregard the results in GSIR which encounter the singularity issue, which only accounts for a very small fraction of all cases. }
The distance correlations between the estimated sufficient predictors $\hat{\f}(\X)$ and the true sufficient predictors $\f(\X)$ are reported in Table \ref{tab:comp-model-d-all}.

\begin{table}[htb]
    \centering
    \small
\begin{tabular}{cccccccc}
\hline
Model & $n$ & BENN-1 & BENN-2 & BENN-1000 & GMDDNet & StoNet & GSIR\\
\hline
\multirow{8}{*}{D-$\rom{4}$}
 & 1000 & 0.45(0.04) & 0.37(0.05) & 0.37(0.04) & 0.46(0.04) & 0.38(0.05) & 0.48(0.03)\\
 & 2000 & 0.52(0.03) & 0.44(0.04) & 0.55(0.03) & 0.56(0.03) & 0.44(0.05) & 0.52(0.02)\\
 & 3000 & 0.55(0.03) & 0.47(0.03) & 0.63(0.03) & 0.60(0.03) & 0.47(0.04) & 0.54(0.03)\\
 & 4000 & 0.58(0.03) & 0.50(0.03) & 0.69(0.02) & 0.62(0.03) & 0.50(0.05) & 0.55(0.02)\\
 & 5000 & 0.58(0.02) & 0.51(0.03) & 0.71(0.02) & 0.62(0.03) & 0.51(0.06) & 0.55(0.02)\\
 & 6000 & 0.60(0.03) & 0.52(0.04) & 0.74(0.02) & 0.63(0.03) & 0.52(0.07) & 0.55(0.02)\\
 & 7000 & 0.60(0.03) & 0.54(0.04) & 0.74(0.03) & 0.64(0.03) & 0.54(0.09) & 0.55(0.02)\\
 & 8000 & 0.61(0.03) & 0.55(0.03) & 0.76(0.03) & 0.64(0.03) & 0.58(0.10) & 0.56(0.02)\\
\hline
\end{tabular}
\caption{Mean (standard deviation) of distance correlations between the estimated sufficient predictors $\hat \f(\X)$ and true ones $\f(\X)$ in Model D-\!\!\rom{4} for BENN, GMDDNet, StoNet and GSIR based on 100 experiments. In the column indices, BENN-$m$ refer to BENN method with the corresponding value of $m$, and the ensemble classes are $\ca G = \{y\}$, $\ca G = \{y, y \hi 2\}$, and $\ca G = \{g(y,y \lo k) = \exp[-(y - y \lo k) \hi 2/(2\sigma \hi 2)], k = 1,\dots,1000\}$ for $m=1,2,1000$, respectively.}
\label{tab:comp-model-d-all}
\end{table}
{From Table \ref{tab:comp-model-d-all} we see that, in Model D-\!\!\rom{4},  BENN-$1000$ performs better than all other methods when the sample size is relatively large. BENN-1000 applies $m=1000$ transformations on $Y$, which does a good job in approximating the conditional distribution of $Y|\X$. The performances of BENN-1 and BENN-2 are somewhat worse than BENN-1000 but are overall comparable with the other estimators.}

In Section \ref{sec:computing-time} of the Supplementary Material, we have also conducted simulations to show that BENN is much faster to compute than the kernel-based nonlinear SDR.

\section{Applications}\label{sec:application}

We have applied BENN   in two data applications. Due to the limited space, we only present {part of} the first application here {and place some additional results in Section \ref{sec:addition-application-1}}. The second application is placed in Section \ref{sec:additional-application}.

\begin{figure}[htbp]
    \centering
    \begin{subfigure}[b]{0.45\textwidth}
        \centering
        \includegraphics[width=\textwidth]{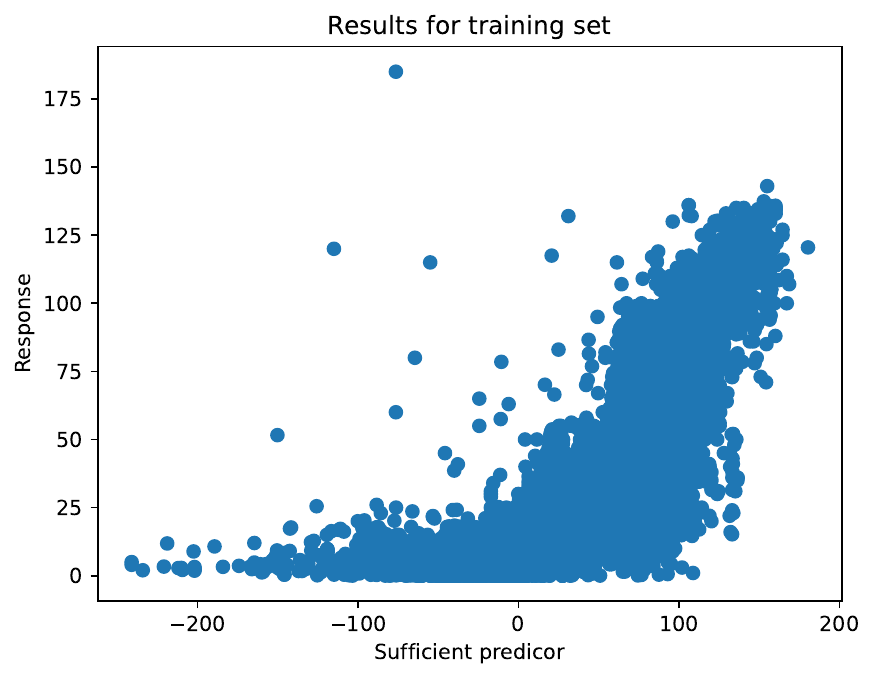} 
                \caption{Results for training set}
    \end{subfigure}
    \hfill
    \begin{subfigure}[b]{0.45\textwidth}
        \centering
        \includegraphics[width=\textwidth]{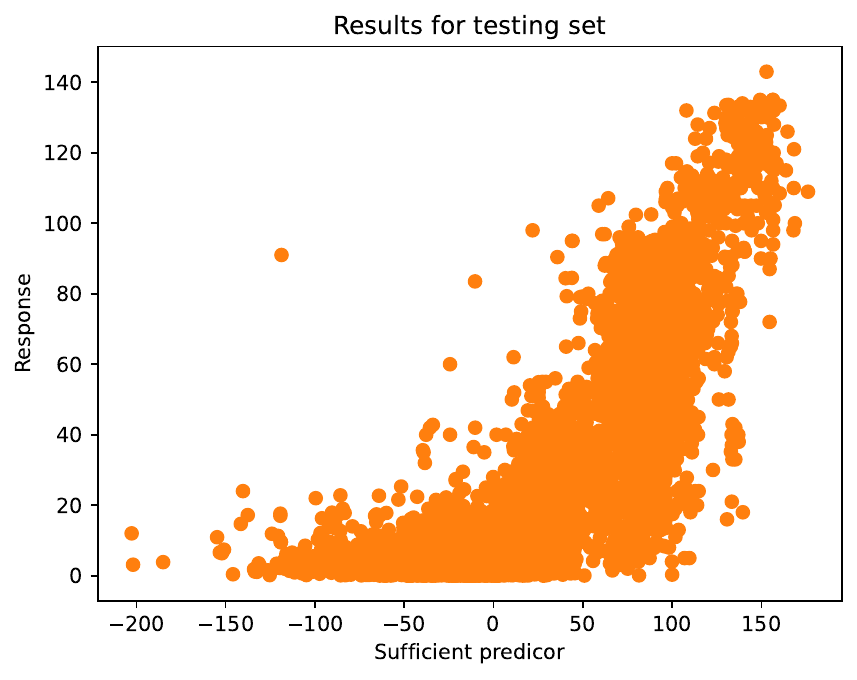} 
                \caption{Results for testing set}
    \end{subfigure}
    \caption{Plots of the sufficient predictors and the original responses in the training and testing sets for the superconductivity dataset using BENN with $m=1$.}
    \label{fig:app-1-benn-1}
\end{figure}

Our first application is a superconductivity dataset, which is studied in \cite{hamidieh2018datadriven}. It contains the critical temperatures of $n=21263$ superconductors as well as their $p=81$ relevant features. The dataset is downloaded from \cite{superconductivity_data_464}.   We assume that the critical temperature is  a function of the 81 relevant features, and  aim to find one sufficient predictor. We first  randomly split the dataset into training and testing sets with proportion 2:1. 
We construct a BENN with parameters  $(81,2,50,1,2,50,m)$, where {$m=1,2,10,50,100,200$,}  and run 400 epochs to get the prediction result. {The cases of $m=1,2,10$ are combined with the ensembles $\ca G = \{y, \cdots, y \hi m \}$, and those of $m=50,100,200$ are combined with the ensembles $\ca G = \{g(y,y \lo k) = \exp[-(y - y \lo k) \hi 2/(2\sigma \hi 2)], k = 1,\dots,m\}$ as in  Section \ref{sec:simulation}. For comparison, we also conduct nonlinear SDR using the GMDDNet, the StoNet with one hidden layer with 50 units, and the  GSIR. For StoNet, to match the scale of the response, we change the default \texttt{sigma\_list}  from $(10 \hi {-3}, 10 \hi {-6})$ to $(10 \hi {-1}, 10 \hi {-2})$.} Then we plot the sufficient predictors and the original responses in the training and testing sets separately as {Figure \ref{fig:app-1-benn-1} below and Figures \ref{fig:app-1-benn-2} through \ref{fig:app-1-gsir} in Section \ref{sec:addition-application-1}. }

{
As can be seen from these plots, BENN with $m=1$ (Figure \ref{fig:app-1-benn-1}) shows a more definite regression pattern than the other three methods: it more compactly wraps around a regression curve. The contrasts are particularly obvious when compared with StoNet (Figure \ref{fig:app-1-stonet}) and GSIR (Figure \ref{fig:app-1-gsir}).} 

{Another advantage of BENN is its fast computing time.} 
The running times are recorded in Table \ref{tab:summary-application-1-all}.   The elapsed times are based on the simulations using 1 node and 8 cores, {either through a Jupyter server or via a batch job running Rscript}. 
{From Table \ref{tab:summary-application-1-all} below, we see that BENN with $m=1$ (which is one of the best performing ensemble sizes) requires substantially shorter computing time than the other methods. 
The computing times for BENN with $m=2,10$ are similar to GMDDNet, faster than the other two methods. 
In particular, GSIR is extremely slow in this case because the sample size is large, which means  GSIR has to invert a large matrix.
}

\section*{Acknowledgement}

We would like to thank two referees and an Associate Editor for their insightful and useful comments and suggestions that have helped us greatly in improving an earlier manuscript. We would also like to thank Professors Faming Liang and Zhou Yu for sharing their computer codes with us.

\bibliography{biblio}

\begin{thebibliography}{}

\bibitem[Bagirov et~al., 2009]{bagirov2009estimation}
Bagirov, A.~M., Clausen, C., and Kohler, M. (2009).
\newblock Estimation of a regression function by maxima of minima of linear
  functions.
\newblock {\em IEEE Transactions on Information Theory}, 55(2):833--845.

\bibitem[Bauer and Kohler, 2019]{bauer2019deep}
Bauer, B. and Kohler, M. (2019).
\newblock {On deep learning as a remedy for the curse of dimensionality in
  nonparametric regression}.
\newblock {\em The Annals of Statistics}, 47(4):2261 -- 2285.

\bibitem[Bourlard and Kamp, 1988]{bourlard1988auto}
Bourlard, H. and Kamp, Y. (1988).
\newblock Auto-association by multilayer perceptrons and singular value
  decomposition.
\newblock {\em Biological Cybernetics}, 59(4):291--294.

\bibitem[Chen et~al., 2024]{chen2024deep}
Chen, Y., Jiao, Y., Qiu, R., and Yu, Z. (2024).
\newblock {Deep nonlinear sufficient dimension reduction}.
\newblock {\em The Annals of Statistics}, 52(3):1201 -- 1226.

\bibitem[Cook and Li, 2004]{cook-li-2004}
Cook, D.~R. and Li, B. (2004).
\newblock Determining the dimension of iterative hessian transformation.
\newblock {\em The Annals of Statistics}, 32:2501--2531.

\bibitem[Cook, 1994]{Cook1994}
Cook, R.~D. (1994).
\newblock Using dimension-reduction subspaces to identify important inputs in
  models of physical systems.
\newblock {\em In 1994 Proceedings of the Section on Physical and Engineering
  Sciences. American Statistical Association, Alexandria, VA.}, pages 18--25.

\bibitem[Cook and Li, 2002]{Cook2002}
Cook, R.~D. and Li, B. (2002).
\newblock {Dimension Reduction for Conditional Mean in Regression}.
\newblock {\em The Annals of Statistics}, 30(2):455--474.

\bibitem[Cook and Weisberg, 1991]{Cook1991}
Cook, R.~D. and Weisberg, S. (1991).
\newblock {Sliced Inverse Regression for Dimension Reduction: Comment}.
\newblock {\em Journal of the American Statistical Association},
  86(414):328--332.

\bibitem[Fertl and Bura, 2022]{fertl2022ensemble}
Fertl, L. and Bura, E. (2022).
\newblock {The ensemble conditional variance estimator for sufficient dimension
  reduction}.
\newblock {\em Electronic Journal of Statistics}, 16(1):1595 -- 1634.

\bibitem[Goodfellow et~al., 2016]{Goodfellow-et-al-2016}
Goodfellow, I., Bengio, Y., and Courville, A. (2016).
\newblock {\em Deep Learning}.
\newblock MIT Press.
\newblock \url{http://www.deeplearningbook.org}.

\bibitem[Gy{\"o}rfi et~al., 2002]{gyorfi2002distribution}
Gy{\"o}rfi, L., Kohler, M., Krzyzak, A., and Walk, H. (2002).
\newblock {\em A Distribution-Free Theory of Nonparametric Regression}.
\newblock Springer Series in Statistics. Springer New York.

\bibitem[Hamidieh, 2018a]{hamidieh2018datadriven}
Hamidieh, K. (2018a).
\newblock A data-driven statistical model for predicting the critical
  temperature of a superconductor.
\newblock {\em Computational Materials Science}, 154:346--354.

\bibitem[Hamidieh, 2018b]{superconductivity_data_464}
Hamidieh, K. (2018b).
\newblock {superconductivity Data}.
\newblock UCI Machine Learning Repository.
\newblock {DOI}: https://doi.org/10.24432/C53P47.

\bibitem[Hinton and Zemel, 1993]{hinton1993autoencoders}
Hinton, G.~E. and Zemel, R. (1993).
\newblock Autoencoders, minimum description length and helmholtz free energy.
\newblock In Cowan, J., Tesauro, G., and Alspector, J., editors, {\em Advances
  in Neural Information Processing Systems}, volume~6. Morgan-Kaufmann.

\bibitem[Huang et~al., 2024]{huang2024deep}
Huang, J., Jiao, Y., Liao, X., Liu, J., and Yu, Z. (2024).
\newblock Deep dimension reduction for supervised representation learning.
\newblock {\em IEEE Transactions on Information Theory}, 70(5):3583--3598.

\bibitem[James et~al., 2023]{james2023introduction}
James, G., Witten, D., Hastie, T., Tibshirani, R., and Taylor, J. (2023).
\newblock {\em An Introduction to Statistical Learning: with Applications in
  Python}.
\newblock Springer International Publishing.

\bibitem[Kapla et~al., 2022]{kapla2022fusing}
Kapla, D., Fertl, L., and Bura, E. (2022).
\newblock Fusing sufficient dimension reduction with neural networks.
\newblock {\em Computational Statistics \& Data Analysis}, 168:107390.

\bibitem[Kohler and Langer, 2021]{kohler2021rate}
Kohler, M. and Langer, S. (2021).
\newblock {On the rate of convergence of fully connected deep neural network
  regression estimates}.
\newblock {\em The Annals of Statistics}, 49(4):2231 -- 2249.

\bibitem[Kramer, 1991]{kramer1991nonlinear}
Kramer, M.~A. (1991).
\newblock Nonlinear principal component analysis using autoassociative neural
  networks.
\newblock {\em AIChE Journal}, 37(2):233--243.

\bibitem[Lee et~al., 2013]{lee2013}
Lee, K.-Y., Li, B., and Chiaromonte, F. (2013).
\newblock A general theory for nonlinear sufficient dimension reduction:
  formulation and estimation.
\newblock {\em The Annals of Statistics}, 41.

\bibitem[Li, 2018]{li2018sufficient}
Li, B. (2018).
\newblock {\em Sufficient Dimension Reduction: Methods and Applications with
  R}.
\newblock Chapman \& Hall/CRC Monographs on Statistics and Applied Probability.
  CRC Press.

\bibitem[Li and Kim, 2024]{li2024sufficient}
Li, B. and Kim, K. (2024).
\newblock On sufficient graphical models.
\newblock {\em Journal of Machine Learning Research}, 25(17):1--64.

\bibitem[Li and Song, 2017]{li2017nonlinear}
Li, B. and Song, J. (2017).
\newblock Nonlinear sufficient dimension reduction for functional data.
\newblock {\em The Annals of Statistics}, pages 1059--1095.

\bibitem[Li et~al., 2008]{li2008projective}
Li, B., Wen, S., and Zhu, L. (2008).
\newblock On a projective resampling method for dimension reduction with
  multivariate responses.
\newblock {\em Journal of the American Statistical Association},
  103(483):1177--1186.

\bibitem[Li, 1991]{Li1991}
Li, K.-C. (1991).
\newblock Sliced inverse regression for dimension reduction.
\newblock {\em Journal of the American Statistical Association},
  86(414):316--327.

\bibitem[Liang et~al., 2022]{liang2022nonlinear}
Liang, S., Sun, Y., and Liang, F. (2022).
\newblock Nonlinear sufficient dimension reduction with a stochastic neural
  network.
\newblock In Oh, A.~H., Agarwal, A., Belgrave, D., and Cho, K., editors, {\em
  Advances in Neural Information Processing Systems}.

\bibitem[Ma and Zhu, 2013]{ma2013review}
Ma, Y. and Zhu, L. (2013).
\newblock A review on dimension reduction.
\newblock {\em International Statistical Review}, 81(1):134--150.

\bibitem[Sch$\ddot{\mbox{o}}$lkopf et~al., 1998]{scholkopf1998nonlinear}
Sch$\ddot{\mbox{o}}$lkopf, B., Smola, A., and M$\ddot{\mbox{u}}$ller, K.~R.
  (1998).
\newblock Nonlinear component analysis as a kernel eigenvalue problem.
\newblock {\em Neural Computation}, 10:1299--1319.

\bibitem[Shen, 2024]{shen2024exploring}
Shen, G. (2024).
\newblock Exploring the complexity of deep neural networks through functional
  equivalence.
\newblock In {\em Forty-first International Conference on Machine Learning}.

\bibitem[Shen et~al., 2020]{shen2020deep}
Shen, Z., Yang, H., and Zhang, S. (2020).
\newblock Deep network approximation characterized by number of neurons.
\newblock {\em Communications in Computational Physics}, 28(5):1768--1811.

\bibitem[Sun and Liang, 2022]{sun2022kernel}
Sun, Y. and Liang, F. (2022).
\newblock {A Kernel-Expanded Stochastic Neural Network}.
\newblock {\em Journal of the Royal Statistical Society Series B: Statistical
  Methodology}, 84(2):547--578.

\bibitem[Sz\'{e}kely et~al., 2007]{szekely2007measuring}
Sz\'{e}kely, G.~J., Rizzo, M.~L., and Bakirov, N.~K. (2007).
\newblock Measuring and testing dependence by correlation of distances.
\newblock {\em The Annals of Statistics}, 35:2769--2794.

\bibitem[Xia et~al., 2002]{xia-tong-li-zhu-2002}
Xia, Y., Tong, H., Li, W.~K., and Zhu, L.-X. (2002).
\newblock An adaptive estimation of dimension reduction space.
\newblock {\em Journal of Royal Statistical Society, Series B}, 64:363--410.

\bibitem[Yin and Cook, 2002]{yin-cook-2002}
Yin, X. and Cook, R.~D. (2002).
\newblock Dimension reduction for the conditional kth moment in regression.
\newblock {\em Journal of the Royal Statistical Society, Series B},
  64:159--175.

\bibitem[Yin and Li, 2011]{yinli2011}
Yin, X. and Li, B. (2011).
\newblock Sufficient dimension reduction based on an ensemble of minimum
  average variance estimators.
\newblock {\em The Annals of Statistics}, 39:3392--3416.

\bibitem[Yuan et~al., 2020]{yuan2020deep}
Yuan, Y., Deng, Y., Zhang, Y., and Qu, A. (2020).
\newblock Deep learning from a statistical perspective.
\newblock {\em Stat}, 9(1):e294.
\newblock e294 sta4.294.

\bibitem[Zeng and Zhu, 2010]{zeng2010integral}
Zeng, P. and Zhu, Y. (2010).
\newblock An integral transform method for estimating the central mean and
  central subspaces.
\newblock {\em Journal of Multivariate Analysis}, 101(1):271--290.

\bibitem[Zhong et~al., 2023]{zhong2023nonlinear}
Zhong, R., Zhang, C., and Zhang, J. (2023).
\newblock Nonlinear functional principal component analysis using neural
  networks.

\bibitem[Zhu and Zeng, 2006]{zhu2006fourier}
Zhu, Y. and Zeng, P. (2006).
\newblock Fourier methods for estimating the central subspace and the central
  mean subspace in regression.
\newblock {\em Journal of the American Statistical Association},
  101(476):1638--1651.

\end{thebibliography}
\end{document}